\documentclass[aps,reprint,superscriptaddress,nofootinbib]{revtex4-1}

\usepackage{graphicx}
\usepackage{amsmath}
\usepackage{amssymb}

\newcommand\blfootnote[1]{%
  \begingroup
  \renewcommand\thefootnote{}\footnote{#1}%
  \addtocounter{footnote}{-1}%
  \endgroup
}

\begin{document}

\title{Tilting the playing field: Dynamical loss functions for machine learning}

\author{Miguel Ruiz-Garc\'ia}

\affiliation{Department of Physics and Astronomy, University of Pennsylvania, Philadelphia, PA 19104, USA}

\affiliation{Department of Applied Mathematics, ETSII, Universidad Polit\'ecnica de Madrid, Madrid, Spain}

\author{Ge Zhang}

\affiliation{Department of Physics and Astronomy, University of Pennsylvania, Philadelphia, PA 19104, USA}

\author{Samuel S. Schoenholz}

\affiliation{Google Brain, Google Inc., Mountain View, California 94043, USA}

\author{Andrea J. Liu}

\affiliation{Department of Physics and Astronomy, University of Pennsylvania, Philadelphia, PA 19104, USA}



\begin{abstract}
We show that learning can be improved by using loss functions that evolve cyclically during training to emphasize one class at a time. In underparameterized networks, such dynamical loss functions can lead to successful training for networks that fail to find deep minima of the standard cross-entropy loss. In overparameterized networks, dynamical loss functions can lead to better generalization. Improvement arises from the interplay of the changing loss landscape with the dynamics of the system as it evolves to minimize the loss. In particular, as the loss function oscillates, instabilities develop in the form of bifurcation cascades, which we study using the Hessian and Neural Tangent Kernel. Valleys in the landscape widen and deepen, and then narrow and rise as the loss landscape changes during a cycle. As the landscape narrows, the learning rate becomes too large and the network becomes unstable and bounces around the valley. This process ultimately pushes the system into deeper and wider regions of the loss landscape and is characterized by decreasing eigenvalues of the Hessian. This results in better regularized models with improved generalization performance.
\end{abstract}

\maketitle


%

\section{Introduction}

\blfootnote{Code reproducing our main results can be found at https://github.com/miguel-rg/dynamical-loss-functions.}

In supervised classification tasks, neural networks learn as they descend a loss function that quantifies their performance. Given a task, there are many components of the learning algorithm that may be tuned to improve performance including: hyperparameters such as the initialization scale~\cite{pmlr-v9-glorot10a,pmlr-v80-xiao18a} or learning rate schedule~\cite{he2016}; the neural network architecture itself~\cite{zoph2016neural}; the types of data augmentation~\cite{cubuk2018autoaugment}; or the optimization algorithm~\cite{Kingma2015AdamAM}. The structure of the loss function also plays an important role in the outcome of learning \cite{choromanska2015loss,soudry2016no,cooper2018loss,verpoort2020archetypal,ballard2017energy,mannelli2019afraid,arous2019landscape}, and it promotes phenomenology reminiscent of physical systems, such as the jamming transition \cite{franz2016simplest,geiger2019jamming,franz2019jamming,franz2019critical,geiger2020scaling,geiger2020perspective}. A possible strategy to improve learning could be to vary the loss function itself; is it possible to tailor the loss function to the training data and to the network architecture to facilitate learning? The first step along this path is to compare how different loss functions perform under the same conditions, see for example \cite{janocha2017loss,rosasco2004loss, kornblith2020s}. However, the plethora of different types of initializations, optimizers, or hyper-parameter combinations, makes it very difficult to find the best option even for a specified set of tasks. Given that choosing an optimal loss function landscape from the beginning is difficult, one might ask if transforming the landscape continuously  during training can lead to a better final result. This takes us to continuation methods, very popular in computational chemistry \cite{stillinger1988nonlinear,wawak1998diffusion,wales1999global}, which had their machine learning counterpart in curriculum learning \cite{bengio2009curriculum}.

Curriculum learning was introduced by Bengio \textit{et al.} \cite{bengio2009curriculum} as a method to improve training of deep neural networks. They found that learning can be improved when the training examples are not randomly presented but are organized in a meaningful order which illustrates gradually more concepts, and gradually more complex ones. In practice, this ``curriculum'' can be achieved by weighing the contribution of easier samples (e.g. most common words) to the loss function more at the beginning and increasing the weight of more difficult samples (e.g. less frequent words) at the end of training. In this way one expects to start with a smoothed-out version of the loss landscape that progressively becomes more complex as training progresses. Since its introduction in 2009, curriculum learning has played a crucial role across deep learning~\cite{pmlr-v48-amodei16, graves2016hybrid,silver2017mastering}. While this approach has been very successful, it often requires additional supervision: for example when labelling images one needs to add a second label for its difficulty.  This requirement can render curriculum learning impractical when there is no clear way to evaluate the difficulty of each training example. 

These considerations raised by curriculum learning suggest new questions: if continuously changing the landscape facilitates learning, why do it only once? Furthermore, training data is already divided into different classes--is it possible to take advantage of this already-existing label for each training example instead of introducing a new label for difficulty? In physical systems, cyclical landscape variation has proven effective in training memory~\cite{keim2011generic,keim2014mechanical,pine2005chaos,hexner2020periodic,sachdeva2020tuning}. In human learning, as well, many educational curricula are developed to expose students to concepts by cycling through them many times rather than learning everything at the same time or learning pieces randomly. Here we extend this approach to neural network training by introducing a \textit{dynamical loss function}. In short, we introduce a time-dependent weight for \textit{each class} to the loss function. During training, the weight applied to each class oscillates, shifting within each cycle to emphasize one class after another. We show in this work that this approach improves training and test accuracy in the underparametrized regime, when the neural network was unable to optimize the standard (static) loss function. Even more surprisingly, it improves test accuracy in the overparameterized regime where the landscape is nearly convex and the final training accuracy is always perfect. Finally, we show how changes in the curvature of the landscape during training lead to bifurcation cascades in the loss function that facilitate better learning.

The advantage of using a dynamical loss function can be understood conceptually as follows. The dynamical loss function changes the loss landscape during minimization, so that although the system is always descending in the instantaneous landscape, it can cross loss-barriers in the static version of the loss function in which each class is weighted equally. The process can be viewed as a sort of peristaltic movement in which the valleys of the landscape alternately sink/grow and rise/shrink, pushing the system into deeper and wider valleys. Progress also occurs when the system falls into valleys that narrow too much for a given learning rate, so that the system caroms from one side of the valley to another, propelling the system into different regions of the landscape. This behavior manifests as bifurcation cascades in the loss function that we will explain in terms of eigenvalues of the Hessian and the Neural Tangent Kernel (NTK)~\cite{jacot2018neural,lee2019wide}. Together, this leads networks trained using dynamical loss functions to move towards wider minima -- a criterion which has been shown to correlate with generalization performance~\cite{zhang2016understanding}.

\section{Myrtle5 and CIFAR10 phase diagrams}

\label{sec_cifar}

\begin{figure*}
    \centering
    \includegraphics[width=0.8\textwidth]{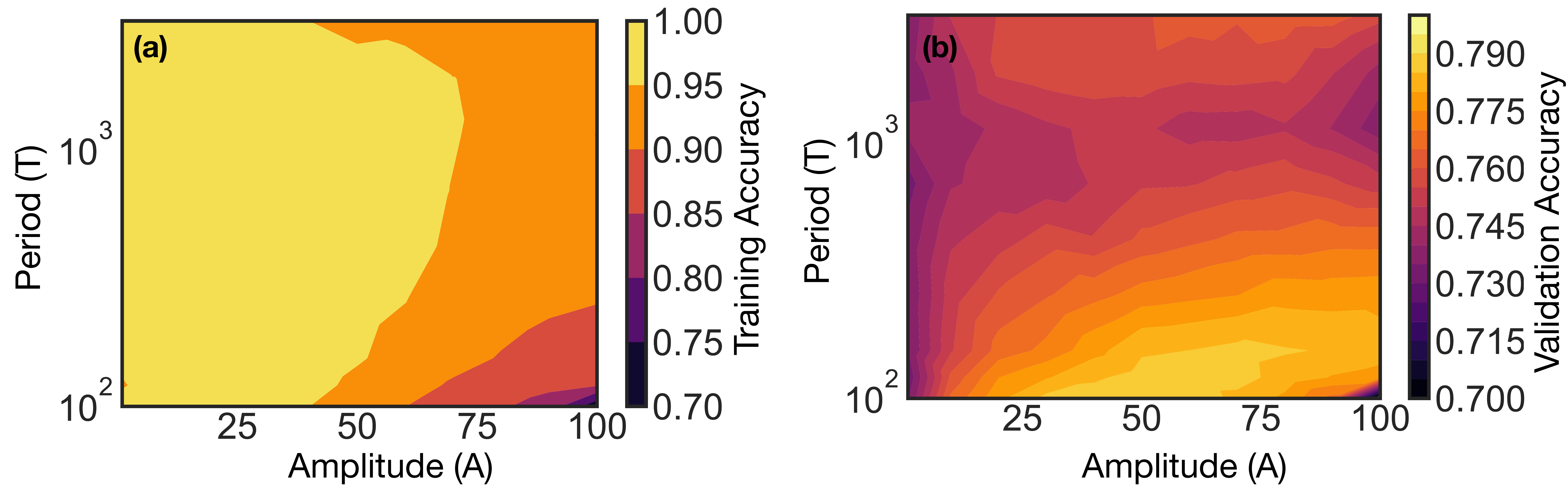}
    \caption{Phase diagrams for the dynamical loss function \eqref{eq_loss} applied to Myrtle5 \cite{shankar2020neural} and CIFAR10. The contour plots represent the training (a) and validation accuracy (b) depending on the amplitude ($A$) and period ($T$) of the oscillations. To create the contour plot we averaged the result of $30$ simulations for each point in a grid in the $(T,A)$ plane. Note that using the standard cross entropy loss function without the oscillations ($\Gamma_i=1$, $A=1$ line in both panels) the system already fitted all the training data (training accuracy $\sim 1$) and achieved a $\sim 0.73$ validation accuracy. However, the validation accuracy improved up to $6\%$ thanks to the oscillations for $A\sim 50$ and $T\sim 100$. This neural network is a realistic setup adapted from \cite{shankar2020neural}. We used $64$ channels, Nesterov optimizer with momentum $=0.9$, minibatch size $512$, a linear learning rate schedule starting at $0$, reaching $0.02$ in the epoch $300$ and decreasing to $0.002$ in the final epoch ($700$). For all $A$ and $T$ the oscillations stopped at epoch $600$ (see the Supplementary Materials for more details).   }
    \label{fig_cifar10}
\end{figure*}

During learning, we denote the number of minimization steps as $t$. We define a dynamical loss function that is a simple variation of cross entropy and changes during learning:
\begin{equation}
    \mathcal{F} = \sum_{j \leq P} \Gamma_{y_j}(t)  \left( - \log \left(\frac{e^{f_{y_j}(x_j,\mathbf{W})}}{\sum_{i}e^{f_i(x_j,\mathbf{W})}} \right) \right)
    \label{eq_loss}
\end{equation}
Where $\Gamma_i$ is a different oscillating factor for \textit{each class $i$}. We further denote $(x_j,y_{j})$ to be an element of the the training set of size $P$, $f(x_j,\mathbf{W})$ is the logit output of the neural network  given a training sample $x_j$ and the value of the trainable parameters $\mathbf{W}$. Here $f(x_j,\mathbf{W})\in\mathbb R^C$ where $C$ is the number of classes. Depending on the values of $\Gamma_i$, the topography of the loss function will change, but the loss function will still vanish at the same global minima, which are unaffected by the value of $\Gamma_i$. This transformation was motivated by recent work in which the topography of the loss function was changed to improve the tuning of physical flow networks \cite{ruiz2019tuning}. Here, we use $\Gamma_i$ to emphasize one class relative to the others for a period $T$, and cycle through all the classes in turn so the total duration of a cycle that passes through all classes is $C T$. To simplify the expression, let us define the time within every period $T$ as
\begin{equation}
   t_T = t\bmod T.
\end{equation}
For simplicity we use a function that linearly increases then decreases with amplitude $A$ so that
\begin{equation}
    g(t_T) = \begin{cases}
    1 + m t_T &\text{for} \quad 0<t_T \le T/2\\
    2A - m t_T - 1  &\text{for} \quad T/2 < t_T \le T
    \end{cases}
\end{equation}
where $A \ge 1$ is the amplitude ($A=1$ corresponds to no oscillations) and $m=2(A-1)/T$. During each period, $\Gamma_i$ increases for one class $i$. We cycle through the classes one by one:
\begin{equation}
    \hat{\Gamma}_i = \begin{cases}
     g(t_T) &\text{for} \quad t/T\bmod C = i\\
    1 &\text{for} \quad  t/T\bmod C  \ne i\\
    \end{cases}
\end{equation}
where $C$ is the number of classes in the dataset. Finally, we normalize these factors,
\begin{equation}
    \Gamma_i = C \frac{ \hat{\Gamma}_i}{\sum_{j=1}^{C} \hat{\Gamma}_j}.
    \label{eq_norm}
\end{equation}
Figure \ref{fig_4} (a) shows the oscillating factors $\Gamma_i$ for the case of a dataset with three classes. Note that due to the normalization, when $\Gamma_i$ increases, $\Gamma_{j\ne i}$ decreases.

To test the effect of oscillations on the outcome of training, we use CIFAR10 as a benchmark, without data augmentation. We train the model $30$ times with the same hyperparameter values to average the results over random initializations of $\mathbf{W}$. We use the Myrtle neural network, introduced in \cite{shankar2020neural}, since it is an efficient convolutional network that achieves good performance with CIFAR10. To obtain enough statistics, we use Myrtle5 with $64$ channels instead of the $1024$ channels used in Ref.~\cite{shankar2020neural}. In all of the experiments we use JAX~\cite{jax2018github} for training, Neural Tangents for computation of the NTK~\cite{neuraltangents2020}, and an open source implementation of the Lanczos algorithm for estimating the spectrum of the Hessian~\cite{ghorbani2019}.

In the standard case without oscillations ($A=1$ in Figure \ref{fig_cifar10}) this model fits all the training data essentially perfectly (training accuracy $\sim 1$) and achieves a modest $\sim 0.73$ validation accuracy. In figure \ref{fig_cifar10} we vary the amplitude $A$ and period $T$ of oscillations to explore the parameter space of the dynamical loss function. We find a region at $25 \lesssim A \lesssim 70$, $T \lesssim 250$ where validation accuracy increases by $\sim 6\%$ to $\sim 0.79$, showing that the dynamical loss function improves generalization significantly.

In addition to the Myrtle5 network, we additionally ran several experiments on a standard Wide Residual Network architecture~\cite{wrn} (see Supplementary Information Sec. IV). Over our limited set of experiments, we did not observe a statistically significant improvement to the test accuracy from using an oscillatory loss. We have several hypotheses for why the oscillatory loss was unhelpful in this case: 1) the oscillatory loss may interact poorly with batch normalization, 2) the network is already well-conditioned and so the oscillations may not lead to further improvements to conditioning, and 3) we used a larger batch size than is typical (1024 vs 128) and trained for only 200 epochs; thus, it might be that the model trained in too few steps to take advantage of the oscillations. It is an interesting avenue for future work to disentangle these effects.  

\section{Understanding the effect of the dynamical loss function in a simpler model}

\subsection{Phase diagrams for the spiral dataset}

\begin{figure}
    \centering
    \includegraphics[width=0.25\textwidth]{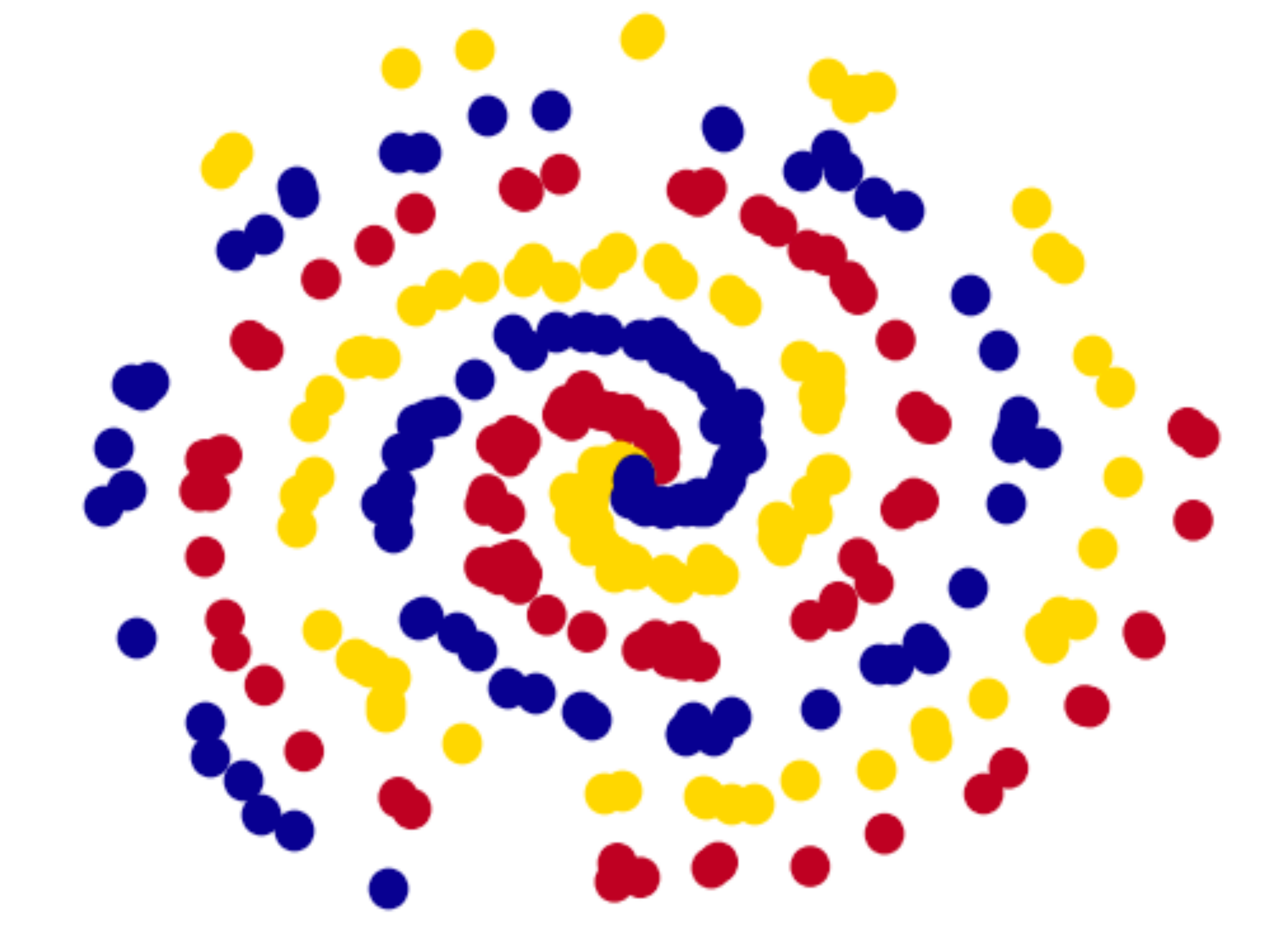}
    \caption{Spiral dataset adapted from \cite{karpathy2020convolutional}. Samples are $2$D arrays belonging to three classes, represented by different colors in the image. Each class follows a different spiral arm plus a small noise.}
    \label{fig_spiral}
\end{figure}

\begin{figure*}
    \centering
    \includegraphics[width=0.7\textwidth]{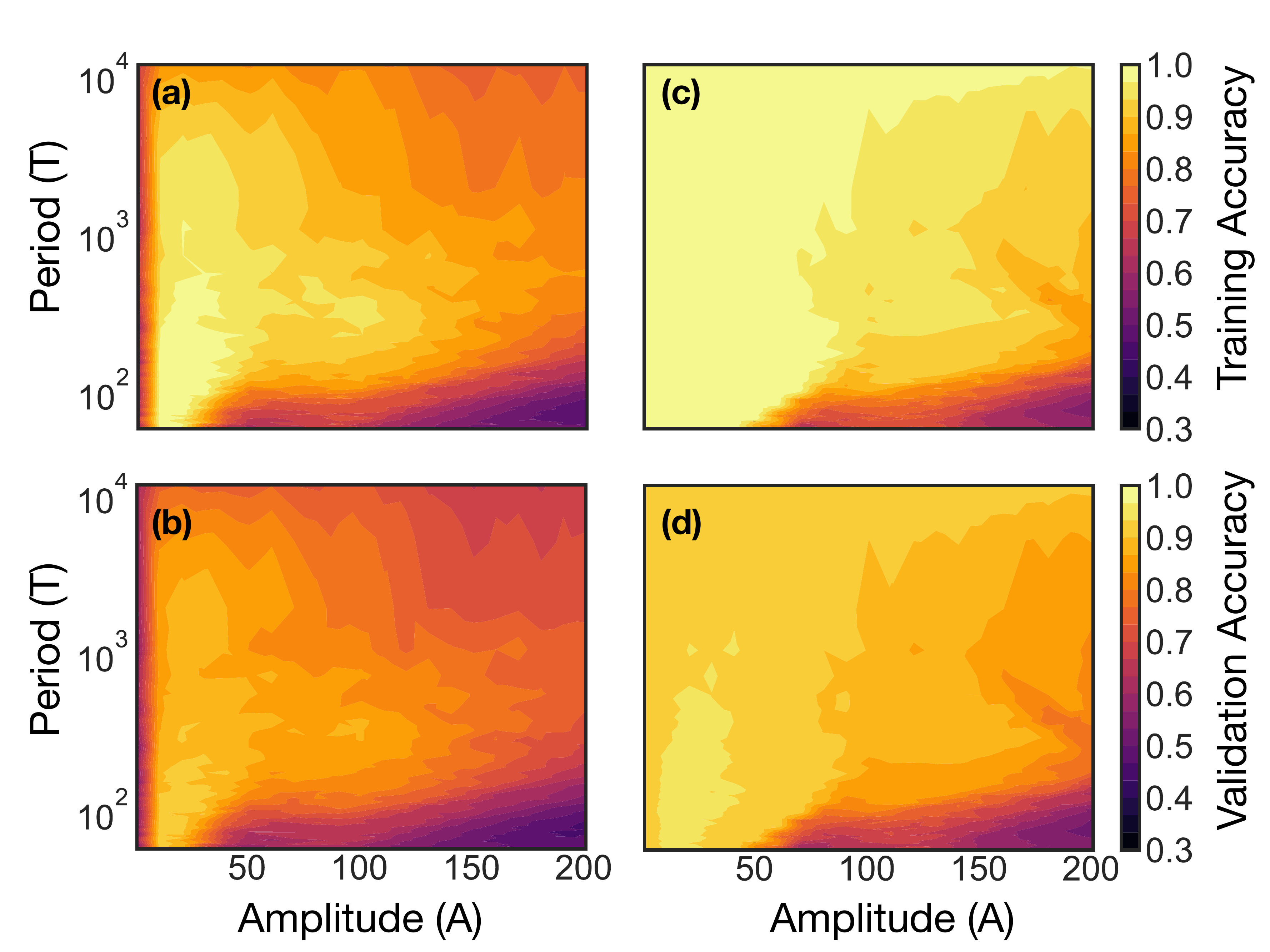}
    \caption{Phase diagrams using the spiral dataset and a neural network with only one hidden layer. We show two examples where the neural network width is $100$ (panels a and b) and $1000$ (panels c and d) respectively. To create the contour plot we averaged the result of $50$ simulations for each point in a grid in the $(T,A)$ plane. In each simulation we used full batch gradient descent for $35000$ steps, a constant learning rate of $1$, and we stopped the oscillations ($\Gamma_i=1$) for the last period. The training dataset is shown in Fig. \ref{fig_spiral} and the validation dataset is analogous to it but with a different distribution of the points along the arms.}
    \label{fig_spiral_PD}
\end{figure*}

To better understand how the oscillations of the dynamical loss function improve generalization, we study a simple but illustrative case. We use synthetic data consisting of points in $2$D that follow a spiral distribution (see Figure \ref{fig_spiral}), with the positions of points belonging to each class following a different arm of the spiral with additional noise (different colors in Fig. \ref{fig_spiral}). In this case we use a neural network with one hidden layer and full batch gradient descent.

Figure~\ref{fig_spiral_PD} shows phase diagrams similar to those in Figure \ref{fig_cifar10}, where we vary the amplitude $A$ and period $T$ of oscillation, for two different network widths, which we will call narrow (100 hidden units) and wide (1000 hidden units), respectively. For the narrow network (left side of Fig.~\ref{fig_spiral_PD}) with the standard cross-entropy loss function (no oscillations; $A=1$) the model is unable to fit the training data, leading to very poor training and validation accuracies ($\sim 0.65$), suggesting that the standard loss function landscape is complex and the network is unable to find a path to a region of low loss. For the dynamical loss function ($A>1$), on the other hand, there is a region in the phase diagram ($5 \lesssim A \lesssim 20$, $T \lesssim 300$), where the training accuracy is nearly perfect and the validation accuracy reaches $\sim 0.9$. Similarly, as we saw in the previous case with Myrtle5 and CIFAR10, when the network is wide enough so that the standard loss landscape is convex (at least in the subspace where training occurs) and the training accuracy is already $\sim 1$ for the standard (static) loss function ($A=1$), there is a regime ($5 \lesssim A \lesssim 25$, $T \lesssim 700$) in which the dynamical loss function improves generalization.

\subsection{Studying the dynamics of learning with a dynamical loss function in terms of the curvature of the landscape}

\begin{figure*}
    \centering
    \includegraphics[width=0.9\textwidth]{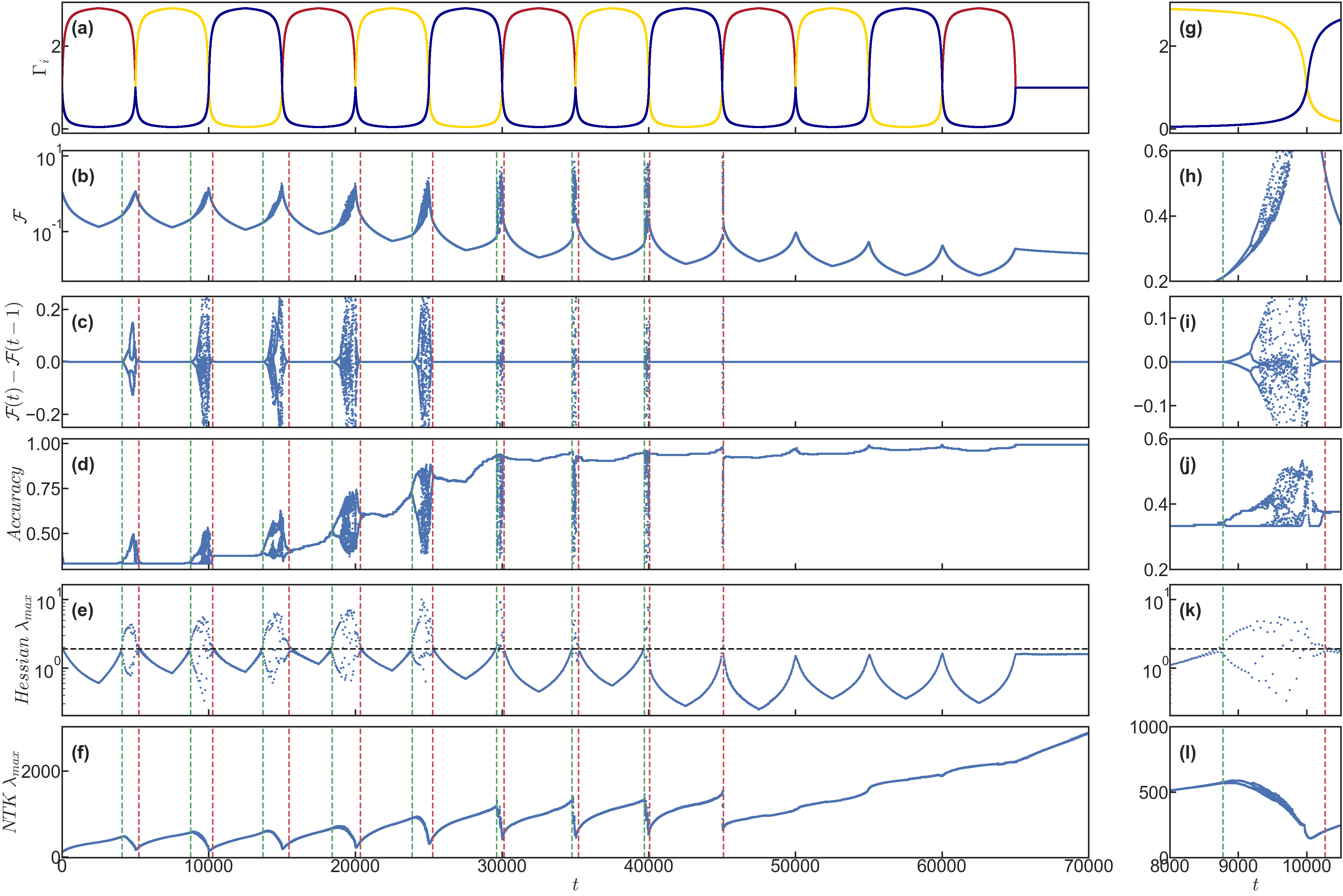}
    \caption{Example of the learning dynamics using a dynamical loss function \eqref{eq_loss}. The width of the hidden layer is $100$. We use $T=5000$ and $A=70$. Panel (a) shows $\Gamma_i$ as training progresses, with colors identifying the corresponding classes shown in figure \ref{fig_spiral}. Panel (b) displays the value of the dynamical loss function $\mathcal{F}(t)$. Panel (c) shows $\mathcal{F}(t)-\mathcal{F}(t-1)$ to display the instabilities more clearly. Panel (d) shows the accuracy of the model during training. Panel (e) shows the largest eigenvalue of the Hessian of the loss function computed using the Lanczos algorithm as described in \cite{ghorbani2019investigation} (we have used an implementation in Google-JAX \cite{gilmer2020}) and panel (f) displays the largest eigenvalue of the NTK \cite{jacot2018neural}. Panels (g-l) correspond to a zoom of panels (a-f) into a region where one bifurcation cascade is present. Vertical green and red dashed lines mark the times at which Hessian $\lambda_{max}(t)-\lambda_{max}(t-1) \sim 0.1$ corresponding to the start and finish of the instabilities. Averaging Hessian $\lambda_{max}$ at these times we get the horizontal dashed line in panel (e), the threshold above which instabilities occur.}
    \label{fig_4}
\end{figure*}

\begin{figure*}
    \centering
    \includegraphics[width=0.9\textwidth]{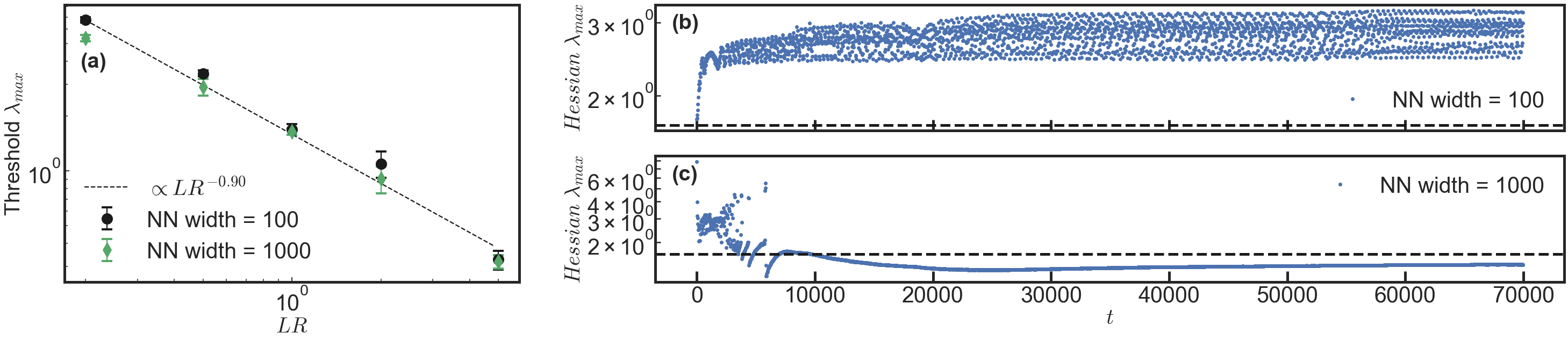}
    \caption{Dependence of the curvature threshold on the learning rate. Panel (a) shows how the threshold of the largest eigenvalue of the Hessian (see Fig. \ref{fig_4}) changes as a function of the learning rate for the two NN widths in Fig. \ref{fig_spiral_PD}. Panel (b) and (c) correspond to two simulations without oscillations ($A=1$) and learning rate $1$. The horizontal dashed line marks the threshold (computed in panel (a)). In panel (b) the system does not find a valley that is wide enough, this prevents learning and $\lambda_{max}$ stays above threshold. Panel (c) presents the same case but with a wider network. The system finds a valley that is wide enough to accommodate the learning rate, the model learns the data and after a transitory it stays below the threshold computed in panel (c) using the dynamical loss function.}
    \label{fig_5}
\end{figure*}

Let us now take a closer look at the training process to understand how loss function oscillations affect learning. During each period $T$, $\Gamma_i>1$ for one class $i$ and  $\Gamma_{j\ne i}<1$. In the most extreme case, $ A \to \infty$, the network only needs to learn class $i$ during that period. For any initial value of the parameters ($\mathbf{W}$) the model can find a solution (the simplest solution is for the network to output the chosen class regardless of input) without making any uphill moves, and therefore the landscape is convex. However, in the next period the network will have to learn a different class, suggesting that the transition between periods will mark the points at which the topography of the landscape becomes more interesting. (Note that right at the transition all $\Gamma_i$ are $1$ and we recover the standard loss function). 

Even without the oscillations ($A=1$), the system does not fully reach a minimum  of the loss function after training--most of the eigenvalues of the Hessian are very small in modulus (even negative) and only a few of the outliers seem to control learning \cite{sagun2017empirical,sagun2016eigenvalues}.
We will refer not to minima but rather to valleys of the loss function, where we consider the projection of parameter space spanned by large-eigenvalue outliers of the Hessian~\cite{gur2018gradient}. 

The behavior of the system as it descends in the dynamical loss function landscape is summarized in Fig.~\ref{fig_4}, for the spiral dataset for a case with a rather high period of $T=5000$ minimization steps and amplitude of $A=70$, chosen for ease of visualization. Panel (a) shows the values of $\Gamma_i$ as learning progresses. Note that due to the normalization \eqref{eq_norm} these values are bounded between $0$ and $3$ (the number of classes $C$). Panel (b) displays the value of the dynamical loss function at each step: in the first half of each period $T$, the loss function decreases as the system descends in a valley, the change of the loss function in each step is small (panel (c)) and the training accuracy is roughly constant (panel (d)). Panel (e) shows the largest eigenvalue of the Hessian, which provides a measure of the width of the valley; during the first half of the period, the system follows a valley that prioritizes learning samples from the chosen class and the largest eigenvalue decreases (the valley widens) as that class is increasingly emphasized.
In this way, the system will move towards a region of parameter space where many (or all) samples belonging to the chosen class are correctly classified. 

It follows that during the first half of the oscillation:
\begin{itemize}
    \item The valley that the system is descending shifts downwards because the network is focusing on learning one class and the contributions to the loss function from the misclassified samples belonging to other classes  contribute less and less as learning progresses ($\Gamma_i(t) > \Gamma_i(t-1)$ and $\Gamma_{j \ne i}(t) < \Gamma_{j \ne i}(t-1)$). 
    \item This valley also widens as other valleys that correctly classify less samples belonging to the chosen class move upwards and shrink (we know about the evolution of the other valleys because it is analogous to the second part of the oscillation, explained below).
\end{itemize}

In the second half of each oscillation $\Gamma_i$ decreases so that the chosen class is now being weighted less and less as time, $t$, progresses. The valley narrows (the largest eigenvalue $\lambda_{max}$ of the Hessian increases; see panel (e)) and rises (the loss function increases even though the system is undergoing gradient descent; see panel (b)). 

To summarize, during the second half of each oscillation, we see the following:
\begin{itemize}
    \item The valley occupied by the system shifts upwards, as the class that the network is focusing on contributes less and less to the loss function and misclassified samples from other classes contribute more ($\Gamma_i(t) < \Gamma_i(t-1)$ and $\Gamma_{j \ne i}(t) > \Gamma_{j \ne i}(t-1)$).
    \item The valley also narrows as other valleys that correctly classify samples belonging to other classes grow and sink (as we saw from the first part of the oscillation). 
\end{itemize}

Additionally, during the second part of each period something remarkable happens when Hessian $\lambda_{max}$ crosses a threshold value, marked by the horizontal dashed line in panel (e). At this time (marked by green dashed lines spanning panels (b-f)), a bifurcation instability appears. Panel (c) and (i) show $\mathcal{F}(t)-\mathcal{F}(t-1)$ where the bifurcation instabilities are clearly visible. As $\Gamma_i \to 1$ there are additional bifurcation instabilities, forming a cascade. 

What is the origin of these bifurcation instabilities? Minimizing the dynamical loss function interweaves two different dynamics: the loss landscape is changing at the same time that the position of the system evolves as it tries to locally minimize the loss function. Thus, both the period $T$ and the learning rate are important. If the learning rate is high enough and the valley is narrow enough, the system is unable to descend the valley and an instability emerges; when other eigenvalues cross this threshold they trigger subsequent bifurcations  creating a cascade. A similar phenomenon is described in detail in~\citet{lewkowycz2020large} where they discuss early learning dynamics with a large learning rate. At the end of the period $T$ (start of the next period) the loss function begins to emphasize another class. Once the system falls into a valley that is favorable to the new class, the valley widens (Hessian $\lambda_{max}$ decreases) and falls below the learning-rate dependent threshold, so the system no longer bounces and begins to smoothly descend the sinking landscape of that valley. In the specific case depicted in Fig.~\ref{fig_4}, the system manages after $10$ oscillations to find a valley that is wide enough so that $\lambda_{max}$ never exceeds threshold during subsequent oscillations. To confirm this hypothesis in the next section we study how the curvature threshold at which instabilities emerge  depends on the learning rate. See also the Supplementary Materials 
for simulations where we plot more than one eigenvalue of the Hessian, and for an example of bifurcating dynamics using Myrtle5 to classify CIFAR10.


\subsection{Dependence of the threshold on the learning rate.}

At each step in the minimization process, the system follows the negative gradient of the loss function, $-\nabla\mathcal{F}$. Taking into account the Taylor series of the loss,
\begin{equation}
    \mathcal{F}(\vec{x}) \sim \mathcal{F}(\vec{a}) + \nabla\mathcal{F}(\vec{a}) (\vec{x}-\vec{a}) + \frac{1}{2} (\vec{x}-\vec{a})^T H\mathcal{F}(\vec{a}) (\vec{x}-\vec{a}),
    \label{eq_taylor}
\end{equation}
where $H\mathcal{F}(\vec{a})$ is the Hessian matrix evaluated at the point $\vec{a}$, our minimization algorithm may fail when the second order terms are of the same order or larger than the first order terms. This is similar to the upper bound for the learning rate using standard loss functions \cite{le1991eigenvalues}. In this case one step in the direction of $-\nabla\mathcal{F}$ can actually take you to a higher value of the loss, as it occurs in the bifurcation cascades. For a learning rate $\eta$, $(\vec{x}-\vec{a}) \propto \eta$, the threshold ($\lambda_{max}^{Th}$) at which the largest eigenvalue of the Hessian makes the first and second order terms in \eqref{eq_taylor} comparable scales as
\begin{equation}
    \lambda_{max}^{Th} \propto \eta^{-1},
    \label{eq_scaling}
\end{equation}
where we have kept only the term corresponding to the largest eigenvalue of the Hessian in \eqref{eq_taylor}. We have also assumed that $-\nabla\mathcal{F}$ is not perpendicular to the eigenvector associated to the largest eigenvalue.

In Fig. \ref{fig_5} (a) we perform simulations equivalent to the one presented in Fig. \ref{fig_4} but with different learning rates. To make the protocol equivalent we rescale the hyperparameters as $T=5000/\eta$ and the total time $70000/\eta$. Panel \ref{fig_5} (a) shows that $\lambda_{max}^{Th}$ does not depend on the network width and it scales as $\sim \eta^{0.9}$, remarkably close to our prediction \eqref{eq_scaling}. Furthermore, although these thresholds are computed using the dynamical loss function, they also control learning in the static loss function (the standard cross entropy). Without oscillations, panel (b) depicts how a NN of width  $100$ cannot learn with $\eta=1$ because its loss function valleys are too narrow  ($\lambda_{max}$ is always above threshold). On the other side, panel (c) shows that after a transitory $\lambda_{max}$ decays below threshold indicating that a wider network produce  wider valleys in the loss function what enables learning with higher learning rates. Note that at least in this case, the underparametrized regime prevents learning because the valleys are too narrow for the learning rate. Even if no bad local minima existed, one may be unable to train the network because an unreasonably small learning rate is necessary.

In figure \ref{fig_4} we observed that when the learning rate is too large for the curvature to accommodate, a instability occurs. The values of Hessian $\lambda_{max}$, ${\cal F}$, ${\cal F}(t)-{\cal F}(t-1)$ and training accuracy bifurcate into two branches that the system visits alternatively in each minimization step. This effect can be understood in terms of the gradient and the Hessian of ${\cal F}$: each minimization step is too long for the curvature so that the system bounces between the walls of the valley. However, in the next section we show that the bifurcation in two branches appears naturally when studying the discrete dynamics of the system using the NTK.

\subsection{Understanding the bifurcations with the NTK}

In addition to the Hessian, the NTK has emerged as a central quantity of interest for understanding the behavior of wide neural networks~\cite{jacot2018neural, lee2019wide, yang2021tensor} whose conditioning has been shown to be closely tied to training and generalization~\cite{xiao2020, dauphin2019, jacot2019}. Even away from the infinite width limit, the NTK can shed light on the dynamics. Figure \ref{fig_4} (f) shows the largest eigenvalue of the NTK. During each of the first 10 oscillations, it increases until the system undergoes a bifurcation instability. It then decreases during the bifurcation cascade. In this section we explain the origin of this phenomenology. We do not develop a rigorous proof but provide the intuition to understand the behavior of the system from the perspective of the NTK.

While the training dynamics in the NTK regime for cross entropy losses have been studied previously~\cite{agarwala2020}, here we find that it suffices to consider the case of the Mean Squared Error (MSE) loss. Note that the arguments in this section closely resemble a previous analysis of neural networks at large learning rates~\cite{lewkowycz2020large}. We write the MSE loss as,
\begin{equation}
    \mathcal{L} = \frac{1}{2 N} \sum_{i,k} \left( f_k(x_i,\mathbf{W}) - y_{i,k}  \right)^2,
\end{equation}
where $(x_i,y_{i,k})$ is the training dataset. Indices $i$ and $k$ are for each sample and class, respectively. $f_k(x_i,\mathbf{W})$ is the output of the neural network (array of dimension $C$) given a training sample $x_i$ and the value of the internal parameters of the NN, $\mathbf{W}=\{w_p\}$. Using gradient descent, the evolution of parameter $w_p$ is
\begin{align}
    \frac{\partial w_p}{\partial t} &= - \eta \frac{\partial \mathcal{L}}{\partial w_p}\nonumber\\ 
    &= -\frac{\eta}{N} \sum_{i,k} \frac{\partial f_k(x_i,\mathbf{W})}{\partial w_p} \left( f_k(x_i,\mathbf{W}) - y_{i,k} \right).
\end{align}
We focus on the evolution of the output of the neural network for an arbitrary sample of the training dataset $x'$,
\begin{align}
     &\frac{\partial f_{k'}(x',\mathbf{W})}{\partial t} = \sum_p \frac{\partial f_{k'}(x',\mathbf{W})}{\partial w_p} \frac{\partial w_p}{\partial t} = \nonumber \\
     &-\frac{\eta}{N} \sum_{i,k,p}
     \frac{\partial f_{k'}(x',\mathbf{W})}{\partial w_p}\frac{\partial f_k(x_i,\mathbf{W})}{\partial w_p} \left( f_k(x_i,\mathbf{W}) - y_{i,k} \right),
     \label{eq_ntk_dyn_1}
\end{align}
and define the NTK as
\begin{equation}
    \Theta_{k',k''}(x',x'') = \sum_p \frac{\partial f_{k'}(x',\mathbf{W})}{\partial w_p}\frac{\partial f_{k''}(x'',\mathbf{W})}{\partial w_p}.
     \label{eq_ntk}
\end{equation}
As an example, in our spiral dataset ($300$ samples and three classes) the NTK can be viewed as a $900$x$900$ matrix. It is useful to consider the difference between the output of the network and the correct label for that sample:
\begin{equation}
    g_k(x) = \left( f_k(x,\mathbf{W})-y_k \right).
    \label{eq_g}
\end{equation}
Combining \eqref{eq_ntk_dyn_1}, \eqref{eq_ntk} and \eqref{eq_g} one obtains
\begin{equation}
    \frac{\partial}{\partial t} g_{k'}(x') = - \frac{\eta}{N} \sum_{i,k} \Theta_{k',k''}(x',x'') g_{k}(x_i).
    \label{eq_ntk_dyn_g}
\end{equation}
In the limit of infinitely small learning rates we can diagonalize the NTK and equation \eqref{eq_ntk_dyn_g} leads to exponential learning of the data, with a rate that is fastest in the directions of the eigenvectors of the NTK associated with the largest eigenvalues. In our case it is more useful to rewrite equation \eqref{eq_ntk_dyn_g} making explicit our discrete dynamics:
\begin{equation}
    g^{t+1}_{k'}(x') = g^{t}_{k'}(x') - \frac{\eta}{N} \sum_{i,k} \Theta^t_{k',k''}(x',x'') g^t_{k}(x_i).
\end{equation}
Diagonalizing the NTK as $\Theta^t_{k',k''}(x',x'') \tilde{g}_j = \lambda_j \tilde{g}_j$ one finds again a stable learning regime when $0<1-\frac{\eta}{N}\lambda_j<1$ (all $\tilde{g}_j$ decrease exponentially). When one NTK eigenvalue increases such that $-1<1-\frac{\eta}{N}\lambda_j<0$, there is still convergence although $\tilde{g}_j$ flips sign with each step. Finally, if one or more eigenvalues cross a threshold such that $1-\frac{\eta}{N}\lambda_j<-1$, learning is unstable and the magnitude of $\tilde{g}_j$ diverges while the sign flips in each step, creating two branches.

Note that in our case the loss function changes during training, and the NTK (as we have defined it here) cannot account for this change, since it only depends on the data and the parameters $\mathbf{W}$. However, we can interpret the instabilities in Figure \ref{fig_4} in terms of the NTK if we take into account the change of the landscape with an \textit{effective} learning rate that changes during training. This agrees with our observations: instabilities emerge in the NTK as bifurcations that change sign between one step and the next one (reminiscent of Fig. \ref{fig_4} (c)); after the first bifurcation starts, the largest eigenvalue of the NTK decreases (Fig. \ref{fig_4} (e)) to try to prevent the divergence. Since the values of the NTK largest eigenvalue (Fig. \ref{fig_4} (e)) are not the same every time that a bifurcation starts (as is the case for the largest eigenvalue of the Hessian) we know that the \textit{effective} learning rate has a non-trivial dependence on the topography of the loss function. 


\section{Discussion}

In this work we have shown that dynamical loss functions can facilitate learning and elucidated the mechanism by which they do so. We have demonstrated that the dynamical loss function can lead to higher training accuracy in the underparametrized region, and can improve the generalization (test or validation accuracy) of overparametrized networks. We have shown this using a realistic model (Myrtle network and CIFAR10) and also using a simple neural network on synthetic data (the spiral dataset). In the latter case we have presented a detailed study of the learning dynamics during gradient descent on the dynamical loss function. In particular, we see that the largest eigenvalue of the Hessian is particularly important in understanding how cycles in the dynamical loss function improve training accuracy by giving rise to bifurcation cascades that allow the system to find wider valleys in the dynamical loss function landscape.

One interesting feature is that the improvement in learning comes in part from using a learning rate that is too fast for the narrowing valley, so that the system bounces instead of descending smoothly within the valley. This feature is somewhat counterintuitive but highly convenient.  Our results show that dynamical loss functions introduce new considerations into the trade-off between speed and accuracy of learning.

In the underparametrized case, learning succeeds with the dynamical loss function while it fails with the standard static loss function for the same values of the hyper-parameters. The attention of most practitioners is focused now on the overparametrized limit, where the model has no problem reaching zero training error. However, for complex problems where the overparametrized limit is infeasible, our results suggest that dynamical loss functions can provide a useful path for learning. 

Learning each class can be considered a different task, so our approach corresponds to switching tasks in a cyclical fashion. The fact that such switching helps learning may seem to be in contradiction with catastrophic forgetting \cite{goodfellow2013empirical,ratcliff1990connectionist,mccloskey1989catastrophic}, where learning new tasks can lead to the forgetting of previous ones.  Figure \ref{fig_spiral_PD} shows that there is an optimum amplitude of the dynamical weighting of the loss function ($A \approx 30$), and that larger values lead to worse performance, in agreement with catastrophic forgetting. Our results show that a strategy that allows learning to proceed on all tasks at all times, but with an oscillating emphasis on one task after another not only avoids catastrophic forgetting but also achieves better results. Our results imply that task-switching schedules should be viewed as a resource for improving learning rather than a liability. Indeed, our results open up a host of interesting questions about how to optimize the choice of static loss function that forms the basis of the dynamical one (e.g. MSE vs. cross entropy) as well as the time-dependence and form of the weighting in the dynamical loss function.

Finally, we note that in the limit $A \to \infty$, the loss function can achieve arbitrarily small values without any uphill moves, starting from any initial weights.
The fact that this limit leads to a convex landscape suggests that it could be interesting to carry out a detailed study of the complexity (number of local minima and saddles of different indices) of the landscape as one varies $A$. This could provide additional valuable insight into the learning process for dynamical loss functions. However, it is probably not enough to simply consider minima and saddles--it is clear from our analysis that valleys play an extremely important role and that their width and depth are important. Studying these topographical features of landscapes as $A$ changes should be very enlightening, although likely very challenging.



\section*{Acknowledgements}

We would like to thank the reviewers for their useful comments and enthusiasm regarding our work. We also thank Stanislav Fort for running some preliminary CIFAR10 experiments. 
This research was supported by the Simons Foundation through the collaboration ``Cracking the Glass Problem'' award $\#$454945 to AJL (MRG,AJL), and Investigator award $\#$327939 (AJL), and by the U.S. Department of Energy, Office of Basic Energy Sciences under Award DE-SC0020963 (GZ). MRG and GZ acknowledge support from the Extreme Science and Engineering Discovery Environment (XSEDE) \cite{towns2014xsede} to use Bridges-2 GPU-AI at the Pittsburgh Supercomputing Center (PSC) through allocation TG-PHY190040. MRG wishes to thank the Istanbul Center for Mathematical Sciences (IMBM) for its hospitality during the workshop on ``Theoretical Advances in Deep Learning''.

 \bibliographystyle{apsrev4-1} 
\bibliography{dyn_loss_bib.bib}

\begin{thebibliography}{60}%
\makeatletter
\providecommand \@ifxundefined [1]{%
 \@ifx{#1\undefined}
}%
\providecommand \@ifnum [1]{%
 \ifnum #1\expandafter \@firstoftwo
 \else \expandafter \@secondoftwo
 \fi
}%
\providecommand \@ifx [1]{%
 \ifx #1\expandafter \@firstoftwo
 \else \expandafter \@secondoftwo
 \fi
}%
\providecommand \natexlab [1]{#1}%
\providecommand \enquote  [1]{``#1''}%
\providecommand \bibnamefont  [1]{#1}%
\providecommand \bibfnamefont [1]{#1}%
\providecommand \citenamefont [1]{#1}%
\providecommand \href@noop [0]{\@secondoftwo}%
\providecommand \href [0]{\begingroup \@sanitize@url \@href}%
\providecommand \@href[1]{\@@startlink{#1}\@@href}%
\providecommand \@@href[1]{\endgroup#1\@@endlink}%
\providecommand \@sanitize@url [0]{\catcode `\\12\catcode `\$12\catcode
  `\&12\catcode `\#12\catcode `\^12\catcode `\_12\catcode `\%12\relax}%
\providecommand \@@startlink[1]{}%
\providecommand \@@endlink[0]{}%
\providecommand \url  [0]{\begingroup\@sanitize@url \@url }%
\providecommand \@url [1]{\endgroup\@href {#1}{\urlprefix }}%
\providecommand \urlprefix  [0]{URL }%
\providecommand \Eprint [0]{\href }%
\providecommand \doibase [0]{http://dx.doi.org/}%
\providecommand \selectlanguage [0]{\@gobble}%
\providecommand \bibinfo  [0]{\@secondoftwo}%
\providecommand \bibfield  [0]{\@secondoftwo}%
\providecommand \translation [1]{[#1]}%
\providecommand \BibitemOpen [0]{}%
\providecommand \bibitemStop [0]{}%
\providecommand \bibitemNoStop [0]{.\EOS\space}%
\providecommand \EOS [0]{\spacefactor3000\relax}%
\providecommand \BibitemShut  [1]{\csname bibitem#1\endcsname}%
\let\auto@bib@innerbib\@empty
\bibitem [{\citenamefont {Glorot}\ and\ \citenamefont
  {Bengio}(2010)}]{pmlr-v9-glorot10a}%
  \BibitemOpen
  \bibfield  {author} {\bibinfo {author} {\bibfnamefont {X.}~\bibnamefont
  {Glorot}}\ and\ \bibinfo {author} {\bibfnamefont {Y.}~\bibnamefont
  {Bengio}},\ }in\ \href@noop {} {\emph {\bibinfo {booktitle} {Proceedings of
  the Thirteenth International Conference on Artificial Intelligence and
  Statistics}}},\ \bibinfo {series} {Proceedings of Machine Learning Research},
  Vol.~\bibinfo {volume} {9},\ \bibinfo {editor} {edited by\ \bibinfo {editor}
  {\bibfnamefont {Y.~W.}\ \bibnamefont {Teh}}\ and\ \bibinfo {editor}
  {\bibfnamefont {M.}~\bibnamefont {Titterington}}}\ (\bibinfo  {publisher}
  {JMLR Workshop and Conference Proceedings},\ \bibinfo {address} {Chia Laguna
  Resort, Sardinia, Italy},\ \bibinfo {year} {2010})\ pp.\ \bibinfo {pages}
  {249--256}\BibitemShut {NoStop}%
\bibitem [{\citenamefont {Xiao}\ \emph {et~al.}(2018)\citenamefont {Xiao},
  \citenamefont {Bahri}, \citenamefont {Sohl-Dickstein}, \citenamefont
  {Schoenholz},\ and\ \citenamefont {Pennington}}]{pmlr-v80-xiao18a}%
  \BibitemOpen
  \bibfield  {author} {\bibinfo {author} {\bibfnamefont {L.}~\bibnamefont
  {Xiao}}, \bibinfo {author} {\bibfnamefont {Y.}~\bibnamefont {Bahri}},
  \bibinfo {author} {\bibfnamefont {J.}~\bibnamefont {Sohl-Dickstein}},
  \bibinfo {author} {\bibfnamefont {S.}~\bibnamefont {Schoenholz}}, \ and\
  \bibinfo {author} {\bibfnamefont {J.}~\bibnamefont {Pennington}},\ }in\
  \href@noop {} {\emph {\bibinfo {booktitle} {Proceedings of the 35th
  International Conference on Machine Learning}}},\ \bibinfo {series}
  {Proceedings of Machine Learning Research}, Vol.~\bibinfo {volume} {80},\
  \bibinfo {editor} {edited by\ \bibinfo {editor} {\bibfnamefont
  {J.}~\bibnamefont {Dy}}\ and\ \bibinfo {editor} {\bibfnamefont
  {A.}~\bibnamefont {Krause}}}\ (\bibinfo  {publisher} {PMLR},\ \bibinfo
  {address} {Stockholmsmässan, Stockholm Sweden},\ \bibinfo {year} {2018})\
  pp.\ \bibinfo {pages} {5393--5402}\BibitemShut {NoStop}%
\bibitem [{\citenamefont {{He}}\ \emph {et~al.}(2016)\citenamefont {{He}},
  \citenamefont {{Zhang}}, \citenamefont {{Ren}},\ and\ \citenamefont
  {{Sun}}}]{he2016}%
  \BibitemOpen
  \bibfield  {author} {\bibinfo {author} {\bibfnamefont {K.}~\bibnamefont
  {{He}}}, \bibinfo {author} {\bibfnamefont {X.}~\bibnamefont {{Zhang}}},
  \bibinfo {author} {\bibfnamefont {S.}~\bibnamefont {{Ren}}}, \ and\ \bibinfo
  {author} {\bibfnamefont {J.}~\bibnamefont {{Sun}}},\ }in\ \href {\doibase
  10.1109/CVPR.2016.90} {\emph {\bibinfo {booktitle} {2016 IEEE Conference on
  Computer Vision and Pattern Recognition (CVPR)}}}\ (\bibinfo {year} {2016})\
  pp.\ \bibinfo {pages} {770--778}\BibitemShut {NoStop}%
\bibitem [{\citenamefont {Zoph}\ and\ \citenamefont
  {Le}(2016)}]{zoph2016neural}%
  \BibitemOpen
  \bibfield  {author} {\bibinfo {author} {\bibfnamefont {B.}~\bibnamefont
  {Zoph}}\ and\ \bibinfo {author} {\bibfnamefont {Q.~V.}\ \bibnamefont {Le}},\
  }\href@noop {} {\bibfield  {journal} {\bibinfo  {journal} {arXiv preprint
  arXiv:1611.01578}\ } (\bibinfo {year} {2016})}\BibitemShut {NoStop}%
\bibitem [{\citenamefont {Cubuk}\ \emph {et~al.}(2018)\citenamefont {Cubuk},
  \citenamefont {Zoph}, \citenamefont {Mane}, \citenamefont {Vasudevan},\ and\
  \citenamefont {Le}}]{cubuk2018autoaugment}%
  \BibitemOpen
  \bibfield  {author} {\bibinfo {author} {\bibfnamefont {E.~D.}\ \bibnamefont
  {Cubuk}}, \bibinfo {author} {\bibfnamefont {B.}~\bibnamefont {Zoph}},
  \bibinfo {author} {\bibfnamefont {D.}~\bibnamefont {Mane}}, \bibinfo {author}
  {\bibfnamefont {V.}~\bibnamefont {Vasudevan}}, \ and\ \bibinfo {author}
  {\bibfnamefont {Q.~V.}\ \bibnamefont {Le}},\ }\href@noop {} {\bibfield
  {journal} {\bibinfo  {journal} {arXiv preprint arXiv:1805.09501}\ } (\bibinfo
  {year} {2018})}\BibitemShut {NoStop}%
\bibitem [{\citenamefont {Kingma}\ and\ \citenamefont
  {Ba}(2015)}]{Kingma2015AdamAM}%
  \BibitemOpen
  \bibfield  {author} {\bibinfo {author} {\bibfnamefont {D.~P.}\ \bibnamefont
  {Kingma}}\ and\ \bibinfo {author} {\bibfnamefont {J.}~\bibnamefont {Ba}},\
  }\href@noop {} {\bibfield  {journal} {\bibinfo  {journal} {CoRR}\ }\textbf
  {\bibinfo {volume} {abs/1412.6980}} (\bibinfo {year} {2015})}\BibitemShut
  {NoStop}%
\bibitem [{\citenamefont {Choromanska}\ \emph {et~al.}(2015)\citenamefont
  {Choromanska}, \citenamefont {Henaff}, \citenamefont {Mathieu}, \citenamefont
  {Arous},\ and\ \citenamefont {LeCun}}]{choromanska2015loss}%
  \BibitemOpen
  \bibfield  {author} {\bibinfo {author} {\bibfnamefont {A.}~\bibnamefont
  {Choromanska}}, \bibinfo {author} {\bibfnamefont {M.}~\bibnamefont {Henaff}},
  \bibinfo {author} {\bibfnamefont {M.}~\bibnamefont {Mathieu}}, \bibinfo
  {author} {\bibfnamefont {G.~B.}\ \bibnamefont {Arous}}, \ and\ \bibinfo
  {author} {\bibfnamefont {Y.}~\bibnamefont {LeCun}},\ }in\ \href@noop {}
  {\emph {\bibinfo {booktitle} {Artificial intelligence and statistics}}}\
  (\bibinfo {organization} {PMLR},\ \bibinfo {year} {2015})\ pp.\ \bibinfo
  {pages} {192--204}\BibitemShut {NoStop}%
\bibitem [{\citenamefont {Soudry}\ and\ \citenamefont
  {Carmon}(2016)}]{soudry2016no}%
  \BibitemOpen
  \bibfield  {author} {\bibinfo {author} {\bibfnamefont {D.}~\bibnamefont
  {Soudry}}\ and\ \bibinfo {author} {\bibfnamefont {Y.}~\bibnamefont
  {Carmon}},\ }\href@noop {} {\bibfield  {journal} {\bibinfo  {journal} {arXiv
  preprint arXiv:1605.08361}\ } (\bibinfo {year} {2016})}\BibitemShut {NoStop}%
\bibitem [{\citenamefont {Cooper}(2018)}]{cooper2018loss}%
  \BibitemOpen
  \bibfield  {author} {\bibinfo {author} {\bibfnamefont {Y.}~\bibnamefont
  {Cooper}},\ }\href@noop {} {\bibfield  {journal} {\bibinfo  {journal} {arXiv
  preprint arXiv:1804.10200}\ } (\bibinfo {year} {2018})}\BibitemShut {NoStop}%
\bibitem [{\citenamefont {Verpoort}\ \emph {et~al.}(2020)\citenamefont
  {Verpoort}, \citenamefont {Wales} \emph {et~al.}}]{verpoort2020archetypal}%
  \BibitemOpen
  \bibfield  {author} {\bibinfo {author} {\bibfnamefont {P.~C.}\ \bibnamefont
  {Verpoort}}, \bibinfo {author} {\bibfnamefont {D.~J.}\ \bibnamefont {Wales}},
   \emph {et~al.},\ }\href@noop {} {\bibfield  {journal} {\bibinfo  {journal}
  {Proceedings of the National Academy of Sciences}\ }\textbf {\bibinfo
  {volume} {117}},\ \bibinfo {pages} {21857} (\bibinfo {year}
  {2020})}\BibitemShut {NoStop}%
\bibitem [{\citenamefont {Ballard}\ \emph {et~al.}(2017)\citenamefont
  {Ballard}, \citenamefont {Das}, \citenamefont {Martiniani}, \citenamefont
  {Mehta}, \citenamefont {Sagun}, \citenamefont {Stevenson},\ and\
  \citenamefont {Wales}}]{ballard2017energy}%
  \BibitemOpen
  \bibfield  {author} {\bibinfo {author} {\bibfnamefont {A.~J.}\ \bibnamefont
  {Ballard}}, \bibinfo {author} {\bibfnamefont {R.}~\bibnamefont {Das}},
  \bibinfo {author} {\bibfnamefont {S.}~\bibnamefont {Martiniani}}, \bibinfo
  {author} {\bibfnamefont {D.}~\bibnamefont {Mehta}}, \bibinfo {author}
  {\bibfnamefont {L.}~\bibnamefont {Sagun}}, \bibinfo {author} {\bibfnamefont
  {J.~D.}\ \bibnamefont {Stevenson}}, \ and\ \bibinfo {author} {\bibfnamefont
  {D.~J.}\ \bibnamefont {Wales}},\ }\href@noop {} {\bibfield  {journal}
  {\bibinfo  {journal} {Physical Chemistry Chemical Physics}\ }\textbf
  {\bibinfo {volume} {19}},\ \bibinfo {pages} {12585} (\bibinfo {year}
  {2017})}\BibitemShut {NoStop}%
\bibitem [{\citenamefont {Mannelli}\ \emph {et~al.}(2019)\citenamefont
  {Mannelli}, \citenamefont {Biroli}, \citenamefont {Cammarota}, \citenamefont
  {Krzakala},\ and\ \citenamefont {Zdeborov{\'a}}}]{mannelli2019afraid}%
  \BibitemOpen
  \bibfield  {author} {\bibinfo {author} {\bibfnamefont {S.~S.}\ \bibnamefont
  {Mannelli}}, \bibinfo {author} {\bibfnamefont {G.}~\bibnamefont {Biroli}},
  \bibinfo {author} {\bibfnamefont {C.}~\bibnamefont {Cammarota}}, \bibinfo
  {author} {\bibfnamefont {F.}~\bibnamefont {Krzakala}}, \ and\ \bibinfo
  {author} {\bibfnamefont {L.}~\bibnamefont {Zdeborov{\'a}}},\ }\href@noop {}
  {\bibfield  {journal} {\bibinfo  {journal} {arXiv preprint arXiv:1907.08226}\
  } (\bibinfo {year} {2019})}\BibitemShut {NoStop}%
\bibitem [{\citenamefont {Arous}\ \emph {et~al.}(2019)\citenamefont {Arous},
  \citenamefont {Mei}, \citenamefont {Montanari},\ and\ \citenamefont
  {Nica}}]{arous2019landscape}%
  \BibitemOpen
  \bibfield  {author} {\bibinfo {author} {\bibfnamefont {G.~B.}\ \bibnamefont
  {Arous}}, \bibinfo {author} {\bibfnamefont {S.}~\bibnamefont {Mei}}, \bibinfo
  {author} {\bibfnamefont {A.}~\bibnamefont {Montanari}}, \ and\ \bibinfo
  {author} {\bibfnamefont {M.}~\bibnamefont {Nica}},\ }\href@noop {} {\bibfield
   {journal} {\bibinfo  {journal} {Communications on Pure and Applied
  Mathematics}\ }\textbf {\bibinfo {volume} {72}},\ \bibinfo {pages} {2282}
  (\bibinfo {year} {2019})}\BibitemShut {NoStop}%
\bibitem [{\citenamefont {Franz}\ and\ \citenamefont
  {Parisi}(2016)}]{franz2016simplest}%
  \BibitemOpen
  \bibfield  {author} {\bibinfo {author} {\bibfnamefont {S.}~\bibnamefont
  {Franz}}\ and\ \bibinfo {author} {\bibfnamefont {G.}~\bibnamefont {Parisi}},\
  }\href@noop {} {\bibfield  {journal} {\bibinfo  {journal} {Journal of Physics
  A: Mathematical and Theoretical}\ }\textbf {\bibinfo {volume} {49}},\
  \bibinfo {pages} {145001} (\bibinfo {year} {2016})}\BibitemShut {NoStop}%
\bibitem [{\citenamefont {Geiger}\ \emph {et~al.}(2019)\citenamefont {Geiger},
  \citenamefont {Spigler}, \citenamefont {d'Ascoli}, \citenamefont {Sagun},
  \citenamefont {Baity-Jesi}, \citenamefont {Biroli},\ and\ \citenamefont
  {Wyart}}]{geiger2019jamming}%
  \BibitemOpen
  \bibfield  {author} {\bibinfo {author} {\bibfnamefont {M.}~\bibnamefont
  {Geiger}}, \bibinfo {author} {\bibfnamefont {S.}~\bibnamefont {Spigler}},
  \bibinfo {author} {\bibfnamefont {S.}~\bibnamefont {d'Ascoli}}, \bibinfo
  {author} {\bibfnamefont {L.}~\bibnamefont {Sagun}}, \bibinfo {author}
  {\bibfnamefont {M.}~\bibnamefont {Baity-Jesi}}, \bibinfo {author}
  {\bibfnamefont {G.}~\bibnamefont {Biroli}}, \ and\ \bibinfo {author}
  {\bibfnamefont {M.}~\bibnamefont {Wyart}},\ }\href@noop {} {\bibfield
  {journal} {\bibinfo  {journal} {Physical Review E}\ }\textbf {\bibinfo
  {volume} {100}},\ \bibinfo {pages} {012115} (\bibinfo {year}
  {2019})}\BibitemShut {NoStop}%
\bibitem [{\citenamefont {Franz}\ \emph
  {et~al.}(2019{\natexlab{a}})\citenamefont {Franz}, \citenamefont {Hwang},\
  and\ \citenamefont {Urbani}}]{franz2019jamming}%
  \BibitemOpen
  \bibfield  {author} {\bibinfo {author} {\bibfnamefont {S.}~\bibnamefont
  {Franz}}, \bibinfo {author} {\bibfnamefont {S.}~\bibnamefont {Hwang}}, \ and\
  \bibinfo {author} {\bibfnamefont {P.}~\bibnamefont {Urbani}},\ }\href@noop {}
  {\bibfield  {journal} {\bibinfo  {journal} {Physical review letters}\
  }\textbf {\bibinfo {volume} {123}},\ \bibinfo {pages} {160602} (\bibinfo
  {year} {2019}{\natexlab{a}})}\BibitemShut {NoStop}%
\bibitem [{\citenamefont {Franz}\ \emph
  {et~al.}(2019{\natexlab{b}})\citenamefont {Franz}, \citenamefont {Sclocchi},\
  and\ \citenamefont {Urbani}}]{franz2019critical}%
  \BibitemOpen
  \bibfield  {author} {\bibinfo {author} {\bibfnamefont {S.}~\bibnamefont
  {Franz}}, \bibinfo {author} {\bibfnamefont {A.}~\bibnamefont {Sclocchi}}, \
  and\ \bibinfo {author} {\bibfnamefont {P.}~\bibnamefont {Urbani}},\
  }\href@noop {} {\bibfield  {journal} {\bibinfo  {journal} {Physical review
  letters}\ }\textbf {\bibinfo {volume} {123}},\ \bibinfo {pages} {115702}
  (\bibinfo {year} {2019}{\natexlab{b}})}\BibitemShut {NoStop}%
\bibitem [{\citenamefont {Geiger}\ \emph
  {et~al.}(2020{\natexlab{a}})\citenamefont {Geiger}, \citenamefont {Jacot},
  \citenamefont {Spigler}, \citenamefont {Gabriel}, \citenamefont {Sagun},
  \citenamefont {d’Ascoli}, \citenamefont {Biroli}, \citenamefont {Hongler},\
  and\ \citenamefont {Wyart}}]{geiger2020scaling}%
  \BibitemOpen
  \bibfield  {author} {\bibinfo {author} {\bibfnamefont {M.}~\bibnamefont
  {Geiger}}, \bibinfo {author} {\bibfnamefont {A.}~\bibnamefont {Jacot}},
  \bibinfo {author} {\bibfnamefont {S.}~\bibnamefont {Spigler}}, \bibinfo
  {author} {\bibfnamefont {F.}~\bibnamefont {Gabriel}}, \bibinfo {author}
  {\bibfnamefont {L.}~\bibnamefont {Sagun}}, \bibinfo {author} {\bibfnamefont
  {S.}~\bibnamefont {d’Ascoli}}, \bibinfo {author} {\bibfnamefont
  {G.}~\bibnamefont {Biroli}}, \bibinfo {author} {\bibfnamefont
  {C.}~\bibnamefont {Hongler}}, \ and\ \bibinfo {author} {\bibfnamefont
  {M.}~\bibnamefont {Wyart}},\ }\href@noop {} {\bibfield  {journal} {\bibinfo
  {journal} {Journal of Statistical Mechanics: Theory and Experiment}\ }\textbf
  {\bibinfo {volume} {2020}},\ \bibinfo {pages} {023401} (\bibinfo {year}
  {2020}{\natexlab{a}})}\BibitemShut {NoStop}%
\bibitem [{\citenamefont {Geiger}\ \emph
  {et~al.}(2020{\natexlab{b}})\citenamefont {Geiger}, \citenamefont {Petrini},\
  and\ \citenamefont {Wyart}}]{geiger2020perspective}%
  \BibitemOpen
  \bibfield  {author} {\bibinfo {author} {\bibfnamefont {M.}~\bibnamefont
  {Geiger}}, \bibinfo {author} {\bibfnamefont {L.}~\bibnamefont {Petrini}}, \
  and\ \bibinfo {author} {\bibfnamefont {M.}~\bibnamefont {Wyart}},\
  }\href@noop {} {\bibfield  {journal} {\bibinfo  {journal} {arXiv preprint
  arXiv:2012.15110}\ } (\bibinfo {year} {2020}{\natexlab{b}})}\BibitemShut
  {NoStop}%
\bibitem [{\citenamefont {Janocha}\ and\ \citenamefont
  {Czarnecki}(2017)}]{janocha2017loss}%
  \BibitemOpen
  \bibfield  {author} {\bibinfo {author} {\bibfnamefont {K.}~\bibnamefont
  {Janocha}}\ and\ \bibinfo {author} {\bibfnamefont {W.~M.}\ \bibnamefont
  {Czarnecki}},\ }\href@noop {} {\bibfield  {journal} {\bibinfo  {journal}
  {arXiv preprint arXiv:1702.05659}\ } (\bibinfo {year} {2017})}\BibitemShut
  {NoStop}%
\bibitem [{\citenamefont {Rosasco}\ \emph {et~al.}(2004)\citenamefont
  {Rosasco}, \citenamefont {Vito}, \citenamefont {Caponnetto}, \citenamefont
  {Piana},\ and\ \citenamefont {Verri}}]{rosasco2004loss}%
  \BibitemOpen
  \bibfield  {author} {\bibinfo {author} {\bibfnamefont {L.}~\bibnamefont
  {Rosasco}}, \bibinfo {author} {\bibfnamefont {E.~D.}\ \bibnamefont {Vito}},
  \bibinfo {author} {\bibfnamefont {A.}~\bibnamefont {Caponnetto}}, \bibinfo
  {author} {\bibfnamefont {M.}~\bibnamefont {Piana}}, \ and\ \bibinfo {author}
  {\bibfnamefont {A.}~\bibnamefont {Verri}},\ }\href@noop {} {\bibfield
  {journal} {\bibinfo  {journal} {Neural Computation}\ }\textbf {\bibinfo
  {volume} {16}},\ \bibinfo {pages} {1063} (\bibinfo {year}
  {2004})}\BibitemShut {NoStop}%
\bibitem [{\citenamefont {Kornblith}\ \emph {et~al.}(2020)\citenamefont
  {Kornblith}, \citenamefont {Lee}, \citenamefont {Chen},\ and\ \citenamefont
  {Norouzi}}]{kornblith2020s}%
  \BibitemOpen
  \bibfield  {author} {\bibinfo {author} {\bibfnamefont {S.}~\bibnamefont
  {Kornblith}}, \bibinfo {author} {\bibfnamefont {H.}~\bibnamefont {Lee}},
  \bibinfo {author} {\bibfnamefont {T.}~\bibnamefont {Chen}}, \ and\ \bibinfo
  {author} {\bibfnamefont {M.}~\bibnamefont {Norouzi}},\ }\href@noop {}
  {\bibfield  {journal} {\bibinfo  {journal} {arXiv preprint arXiv:2010.16402}\
  } (\bibinfo {year} {2020})}\BibitemShut {NoStop}%
\bibitem [{\citenamefont {Stillinger}\ and\ \citenamefont
  {Weber}(1988)}]{stillinger1988nonlinear}%
  \BibitemOpen
  \bibfield  {author} {\bibinfo {author} {\bibfnamefont {F.}~\bibnamefont
  {Stillinger}}\ and\ \bibinfo {author} {\bibfnamefont {T.}~\bibnamefont
  {Weber}},\ }\href@noop {} {\bibfield  {journal} {\bibinfo  {journal} {Journal
  of statistical physics}\ }\textbf {\bibinfo {volume} {52}},\ \bibinfo {pages}
  {1429} (\bibinfo {year} {1988})}\BibitemShut {NoStop}%
\bibitem [{\citenamefont {Wawak}\ \emph {et~al.}(1998)\citenamefont {Wawak},
  \citenamefont {Pillardy}, \citenamefont {Liwo}, \citenamefont {Gibson},\ and\
  \citenamefont {Scheraga}}]{wawak1998diffusion}%
  \BibitemOpen
  \bibfield  {author} {\bibinfo {author} {\bibfnamefont {R.~J.}\ \bibnamefont
  {Wawak}}, \bibinfo {author} {\bibfnamefont {J.}~\bibnamefont {Pillardy}},
  \bibinfo {author} {\bibfnamefont {A.}~\bibnamefont {Liwo}}, \bibinfo {author}
  {\bibfnamefont {K.~D.}\ \bibnamefont {Gibson}}, \ and\ \bibinfo {author}
  {\bibfnamefont {H.~A.}\ \bibnamefont {Scheraga}},\ }\href@noop {} {\bibfield
  {journal} {\bibinfo  {journal} {The Journal of Physical Chemistry A}\
  }\textbf {\bibinfo {volume} {102}},\ \bibinfo {pages} {2904} (\bibinfo {year}
  {1998})}\BibitemShut {NoStop}%
\bibitem [{\citenamefont {Wales}\ and\ \citenamefont
  {Scheraga}(1999)}]{wales1999global}%
  \BibitemOpen
  \bibfield  {author} {\bibinfo {author} {\bibfnamefont {D.~J.}\ \bibnamefont
  {Wales}}\ and\ \bibinfo {author} {\bibfnamefont {H.~A.}\ \bibnamefont
  {Scheraga}},\ }\href@noop {} {\bibfield  {journal} {\bibinfo  {journal}
  {Science}\ }\textbf {\bibinfo {volume} {285}},\ \bibinfo {pages} {1368}
  (\bibinfo {year} {1999})}\BibitemShut {NoStop}%
\bibitem [{\citenamefont {Bengio}\ \emph {et~al.}(2009)\citenamefont {Bengio},
  \citenamefont {Louradour}, \citenamefont {Collobert},\ and\ \citenamefont
  {Weston}}]{bengio2009curriculum}%
  \BibitemOpen
  \bibfield  {author} {\bibinfo {author} {\bibfnamefont {Y.}~\bibnamefont
  {Bengio}}, \bibinfo {author} {\bibfnamefont {J.}~\bibnamefont {Louradour}},
  \bibinfo {author} {\bibfnamefont {R.}~\bibnamefont {Collobert}}, \ and\
  \bibinfo {author} {\bibfnamefont {J.}~\bibnamefont {Weston}},\ }in\
  \href@noop {} {\emph {\bibinfo {booktitle} {Proceedings of the 26th annual
  international conference on machine learning}}}\ (\bibinfo {year} {2009})\
  pp.\ \bibinfo {pages} {41--48}\BibitemShut {NoStop}%
\bibitem [{\citenamefont {Amodei}\ \emph {et~al.}(2016)\citenamefont {Amodei},
  \citenamefont {Ananthanarayanan}, \citenamefont {Anubhai}, \citenamefont
  {Bai}, \citenamefont {Battenberg}, \citenamefont {Case}, \citenamefont
  {Casper}, \citenamefont {Catanzaro}, \citenamefont {Cheng}, \citenamefont
  {Chen}, \citenamefont {Chen}, \citenamefont {Chen}, \citenamefont {Chen},
  \citenamefont {Chrzanowski}, \citenamefont {Coates}, \citenamefont {Diamos},
  \citenamefont {Ding}, \citenamefont {Du}, \citenamefont {Elsen},
  \citenamefont {Engel}, \citenamefont {Fang}, \citenamefont {Fan},
  \citenamefont {Fougner}, \citenamefont {Gao}, \citenamefont {Gong},
  \citenamefont {Hannun}, \citenamefont {Han}, \citenamefont {Johannes},
  \citenamefont {Jiang}, \citenamefont {Ju}, \citenamefont {Jun}, \citenamefont
  {LeGresley}, \citenamefont {Lin}, \citenamefont {Liu}, \citenamefont {Liu},
  \citenamefont {Li}, \citenamefont {Li}, \citenamefont {Ma}, \citenamefont
  {Narang}, \citenamefont {Ng}, \citenamefont {Ozair}, \citenamefont {Peng},
  \citenamefont {Prenger}, \citenamefont {Qian}, \citenamefont {Quan},
  \citenamefont {Raiman}, \citenamefont {Rao}, \citenamefont {Satheesh},
  \citenamefont {Seetapun}, \citenamefont {Sengupta}, \citenamefont {Srinet},
  \citenamefont {Sriram}, \citenamefont {Tang}, \citenamefont {Tang},
  \citenamefont {Wang}, \citenamefont {Wang}, \citenamefont {Wang},
  \citenamefont {Wang}, \citenamefont {Wang}, \citenamefont {Wang},
  \citenamefont {Wu}, \citenamefont {Wei}, \citenamefont {Xiao}, \citenamefont
  {Xie}, \citenamefont {Xie}, \citenamefont {Yogatama}, \citenamefont {Yuan},
  \citenamefont {Zhan},\ and\ \citenamefont {Zhu}}]{pmlr-v48-amodei16}%
  \BibitemOpen
  \bibfield  {author} {\bibinfo {author} {\bibfnamefont {D.}~\bibnamefont
  {Amodei}}, \bibinfo {author} {\bibfnamefont {S.}~\bibnamefont
  {Ananthanarayanan}}, \bibinfo {author} {\bibfnamefont {R.}~\bibnamefont
  {Anubhai}}, \bibinfo {author} {\bibfnamefont {J.}~\bibnamefont {Bai}},
  \bibinfo {author} {\bibfnamefont {E.}~\bibnamefont {Battenberg}}, \bibinfo
  {author} {\bibfnamefont {C.}~\bibnamefont {Case}}, \bibinfo {author}
  {\bibfnamefont {J.}~\bibnamefont {Casper}}, \bibinfo {author} {\bibfnamefont
  {B.}~\bibnamefont {Catanzaro}}, \bibinfo {author} {\bibfnamefont
  {Q.}~\bibnamefont {Cheng}}, \bibinfo {author} {\bibfnamefont
  {G.}~\bibnamefont {Chen}}, \bibinfo {author} {\bibfnamefont {J.}~\bibnamefont
  {Chen}}, \bibinfo {author} {\bibfnamefont {J.}~\bibnamefont {Chen}}, \bibinfo
  {author} {\bibfnamefont {Z.}~\bibnamefont {Chen}}, \bibinfo {author}
  {\bibfnamefont {M.}~\bibnamefont {Chrzanowski}}, \bibinfo {author}
  {\bibfnamefont {A.}~\bibnamefont {Coates}}, \bibinfo {author} {\bibfnamefont
  {G.}~\bibnamefont {Diamos}}, \bibinfo {author} {\bibfnamefont
  {K.}~\bibnamefont {Ding}}, \bibinfo {author} {\bibfnamefont {N.}~\bibnamefont
  {Du}}, \bibinfo {author} {\bibfnamefont {E.}~\bibnamefont {Elsen}}, \bibinfo
  {author} {\bibfnamefont {J.}~\bibnamefont {Engel}}, \bibinfo {author}
  {\bibfnamefont {W.}~\bibnamefont {Fang}}, \bibinfo {author} {\bibfnamefont
  {L.}~\bibnamefont {Fan}}, \bibinfo {author} {\bibfnamefont {C.}~\bibnamefont
  {Fougner}}, \bibinfo {author} {\bibfnamefont {L.}~\bibnamefont {Gao}},
  \bibinfo {author} {\bibfnamefont {C.}~\bibnamefont {Gong}}, \bibinfo {author}
  {\bibfnamefont {A.}~\bibnamefont {Hannun}}, \bibinfo {author} {\bibfnamefont
  {T.}~\bibnamefont {Han}}, \bibinfo {author} {\bibfnamefont {L.}~\bibnamefont
  {Johannes}}, \bibinfo {author} {\bibfnamefont {B.}~\bibnamefont {Jiang}},
  \bibinfo {author} {\bibfnamefont {C.}~\bibnamefont {Ju}}, \bibinfo {author}
  {\bibfnamefont {B.}~\bibnamefont {Jun}}, \bibinfo {author} {\bibfnamefont
  {P.}~\bibnamefont {LeGresley}}, \bibinfo {author} {\bibfnamefont
  {L.}~\bibnamefont {Lin}}, \bibinfo {author} {\bibfnamefont {J.}~\bibnamefont
  {Liu}}, \bibinfo {author} {\bibfnamefont {Y.}~\bibnamefont {Liu}}, \bibinfo
  {author} {\bibfnamefont {W.}~\bibnamefont {Li}}, \bibinfo {author}
  {\bibfnamefont {X.}~\bibnamefont {Li}}, \bibinfo {author} {\bibfnamefont
  {D.}~\bibnamefont {Ma}}, \bibinfo {author} {\bibfnamefont {S.}~\bibnamefont
  {Narang}}, \bibinfo {author} {\bibfnamefont {A.}~\bibnamefont {Ng}}, \bibinfo
  {author} {\bibfnamefont {S.}~\bibnamefont {Ozair}}, \bibinfo {author}
  {\bibfnamefont {Y.}~\bibnamefont {Peng}}, \bibinfo {author} {\bibfnamefont
  {R.}~\bibnamefont {Prenger}}, \bibinfo {author} {\bibfnamefont
  {S.}~\bibnamefont {Qian}}, \bibinfo {author} {\bibfnamefont {Z.}~\bibnamefont
  {Quan}}, \bibinfo {author} {\bibfnamefont {J.}~\bibnamefont {Raiman}},
  \bibinfo {author} {\bibfnamefont {V.}~\bibnamefont {Rao}}, \bibinfo {author}
  {\bibfnamefont {S.}~\bibnamefont {Satheesh}}, \bibinfo {author}
  {\bibfnamefont {D.}~\bibnamefont {Seetapun}}, \bibinfo {author}
  {\bibfnamefont {S.}~\bibnamefont {Sengupta}}, \bibinfo {author}
  {\bibfnamefont {K.}~\bibnamefont {Srinet}}, \bibinfo {author} {\bibfnamefont
  {A.}~\bibnamefont {Sriram}}, \bibinfo {author} {\bibfnamefont
  {H.}~\bibnamefont {Tang}}, \bibinfo {author} {\bibfnamefont {L.}~\bibnamefont
  {Tang}}, \bibinfo {author} {\bibfnamefont {C.}~\bibnamefont {Wang}}, \bibinfo
  {author} {\bibfnamefont {J.}~\bibnamefont {Wang}}, \bibinfo {author}
  {\bibfnamefont {K.}~\bibnamefont {Wang}}, \bibinfo {author} {\bibfnamefont
  {Y.}~\bibnamefont {Wang}}, \bibinfo {author} {\bibfnamefont {Z.}~\bibnamefont
  {Wang}}, \bibinfo {author} {\bibfnamefont {Z.}~\bibnamefont {Wang}}, \bibinfo
  {author} {\bibfnamefont {S.}~\bibnamefont {Wu}}, \bibinfo {author}
  {\bibfnamefont {L.}~\bibnamefont {Wei}}, \bibinfo {author} {\bibfnamefont
  {B.}~\bibnamefont {Xiao}}, \bibinfo {author} {\bibfnamefont {W.}~\bibnamefont
  {Xie}}, \bibinfo {author} {\bibfnamefont {Y.}~\bibnamefont {Xie}}, \bibinfo
  {author} {\bibfnamefont {D.}~\bibnamefont {Yogatama}}, \bibinfo {author}
  {\bibfnamefont {B.}~\bibnamefont {Yuan}}, \bibinfo {author} {\bibfnamefont
  {J.}~\bibnamefont {Zhan}}, \ and\ \bibinfo {author} {\bibfnamefont
  {Z.}~\bibnamefont {Zhu}},\ }in\ \href@noop {} {\emph {\bibinfo {booktitle}
  {Proceedings of The 33rd International Conference on Machine Learning}}},\
  \bibinfo {series} {Proceedings of Machine Learning Research}, Vol.~\bibinfo
  {volume} {48},\ \bibinfo {editor} {edited by\ \bibinfo {editor}
  {\bibfnamefont {M.~F.}\ \bibnamefont {Balcan}}\ and\ \bibinfo {editor}
  {\bibfnamefont {K.~Q.}\ \bibnamefont {Weinberger}}}\ (\bibinfo  {publisher}
  {PMLR},\ \bibinfo {address} {New York, New York, USA},\ \bibinfo {year}
  {2016})\ pp.\ \bibinfo {pages} {173--182}\BibitemShut {NoStop}%
\bibitem [{\citenamefont {Graves}\ \emph {et~al.}(2016)\citenamefont {Graves},
  \citenamefont {Wayne}, \citenamefont {Reynolds}, \citenamefont {Harley},
  \citenamefont {Danihelka}, \citenamefont {Grabska-Barwi{\'n}ska},
  \citenamefont {Colmenarejo}, \citenamefont {Grefenstette}, \citenamefont
  {Ramalho}, \citenamefont {Agapiou} \emph {et~al.}}]{graves2016hybrid}%
  \BibitemOpen
  \bibfield  {author} {\bibinfo {author} {\bibfnamefont {A.}~\bibnamefont
  {Graves}}, \bibinfo {author} {\bibfnamefont {G.}~\bibnamefont {Wayne}},
  \bibinfo {author} {\bibfnamefont {M.}~\bibnamefont {Reynolds}}, \bibinfo
  {author} {\bibfnamefont {T.}~\bibnamefont {Harley}}, \bibinfo {author}
  {\bibfnamefont {I.}~\bibnamefont {Danihelka}}, \bibinfo {author}
  {\bibfnamefont {A.}~\bibnamefont {Grabska-Barwi{\'n}ska}}, \bibinfo {author}
  {\bibfnamefont {S.~G.}\ \bibnamefont {Colmenarejo}}, \bibinfo {author}
  {\bibfnamefont {E.}~\bibnamefont {Grefenstette}}, \bibinfo {author}
  {\bibfnamefont {T.}~\bibnamefont {Ramalho}}, \bibinfo {author} {\bibfnamefont
  {J.}~\bibnamefont {Agapiou}},  \emph {et~al.},\ }\href@noop {} {\bibfield
  {journal} {\bibinfo  {journal} {Nature}\ }\textbf {\bibinfo {volume} {538}},\
  \bibinfo {pages} {471} (\bibinfo {year} {2016})}\BibitemShut {NoStop}%
\bibitem [{\citenamefont {Silver}\ \emph {et~al.}(2017)\citenamefont {Silver},
  \citenamefont {Schrittwieser}, \citenamefont {Simonyan}, \citenamefont
  {Antonoglou}, \citenamefont {Huang}, \citenamefont {Guez}, \citenamefont
  {Hubert}, \citenamefont {Baker}, \citenamefont {Lai}, \citenamefont {Bolton}
  \emph {et~al.}}]{silver2017mastering}%
  \BibitemOpen
  \bibfield  {author} {\bibinfo {author} {\bibfnamefont {D.}~\bibnamefont
  {Silver}}, \bibinfo {author} {\bibfnamefont {J.}~\bibnamefont
  {Schrittwieser}}, \bibinfo {author} {\bibfnamefont {K.}~\bibnamefont
  {Simonyan}}, \bibinfo {author} {\bibfnamefont {I.}~\bibnamefont
  {Antonoglou}}, \bibinfo {author} {\bibfnamefont {A.}~\bibnamefont {Huang}},
  \bibinfo {author} {\bibfnamefont {A.}~\bibnamefont {Guez}}, \bibinfo {author}
  {\bibfnamefont {T.}~\bibnamefont {Hubert}}, \bibinfo {author} {\bibfnamefont
  {L.}~\bibnamefont {Baker}}, \bibinfo {author} {\bibfnamefont
  {M.}~\bibnamefont {Lai}}, \bibinfo {author} {\bibfnamefont {A.}~\bibnamefont
  {Bolton}},  \emph {et~al.},\ }\href@noop {} {\bibfield  {journal} {\bibinfo
  {journal} {nature}\ }\textbf {\bibinfo {volume} {550}},\ \bibinfo {pages}
  {354} (\bibinfo {year} {2017})}\BibitemShut {NoStop}%
\bibitem [{\citenamefont {Keim}\ and\ \citenamefont
  {Nagel}(2011)}]{keim2011generic}%
  \BibitemOpen
  \bibfield  {author} {\bibinfo {author} {\bibfnamefont {N.~C.}\ \bibnamefont
  {Keim}}\ and\ \bibinfo {author} {\bibfnamefont {S.~R.}\ \bibnamefont
  {Nagel}},\ }\href@noop {} {\bibfield  {journal} {\bibinfo  {journal}
  {Physical review letters}\ }\textbf {\bibinfo {volume} {107}},\ \bibinfo
  {pages} {010603} (\bibinfo {year} {2011})}\BibitemShut {NoStop}%
\bibitem [{\citenamefont {Keim}\ and\ \citenamefont
  {Arratia}(2014)}]{keim2014mechanical}%
  \BibitemOpen
  \bibfield  {author} {\bibinfo {author} {\bibfnamefont {N.~C.}\ \bibnamefont
  {Keim}}\ and\ \bibinfo {author} {\bibfnamefont {P.~E.}\ \bibnamefont
  {Arratia}},\ }\href@noop {} {\bibfield  {journal} {\bibinfo  {journal}
  {Physical review letters}\ }\textbf {\bibinfo {volume} {112}},\ \bibinfo
  {pages} {028302} (\bibinfo {year} {2014})}\BibitemShut {NoStop}%
\bibitem [{\citenamefont {Pine}\ \emph {et~al.}(2005)\citenamefont {Pine},
  \citenamefont {Gollub}, \citenamefont {Brady},\ and\ \citenamefont
  {Leshansky}}]{pine2005chaos}%
  \BibitemOpen
  \bibfield  {author} {\bibinfo {author} {\bibfnamefont {D.~J.}\ \bibnamefont
  {Pine}}, \bibinfo {author} {\bibfnamefont {J.~P.}\ \bibnamefont {Gollub}},
  \bibinfo {author} {\bibfnamefont {J.~F.}\ \bibnamefont {Brady}}, \ and\
  \bibinfo {author} {\bibfnamefont {A.~M.}\ \bibnamefont {Leshansky}},\
  }\href@noop {} {\bibfield  {journal} {\bibinfo  {journal} {Nature}\ }\textbf
  {\bibinfo {volume} {438}},\ \bibinfo {pages} {997} (\bibinfo {year}
  {2005})}\BibitemShut {NoStop}%
\bibitem [{\citenamefont {Hexner}\ \emph {et~al.}(2020)\citenamefont {Hexner},
  \citenamefont {Liu},\ and\ \citenamefont {Nagel}}]{hexner2020periodic}%
  \BibitemOpen
  \bibfield  {author} {\bibinfo {author} {\bibfnamefont {D.}~\bibnamefont
  {Hexner}}, \bibinfo {author} {\bibfnamefont {A.~J.}\ \bibnamefont {Liu}}, \
  and\ \bibinfo {author} {\bibfnamefont {S.~R.}\ \bibnamefont {Nagel}},\
  }\href@noop {} {\bibfield  {journal} {\bibinfo  {journal} {Proceedings of the
  National Academy of Sciences}\ }\textbf {\bibinfo {volume} {117}},\ \bibinfo
  {pages} {31690} (\bibinfo {year} {2020})}\BibitemShut {NoStop}%
\bibitem [{\citenamefont {Sachdeva}\ \emph {et~al.}(2020)\citenamefont
  {Sachdeva}, \citenamefont {Husain}, \citenamefont {Sheng}, \citenamefont
  {Wang},\ and\ \citenamefont {Murugan}}]{sachdeva2020tuning}%
  \BibitemOpen
  \bibfield  {author} {\bibinfo {author} {\bibfnamefont {V.}~\bibnamefont
  {Sachdeva}}, \bibinfo {author} {\bibfnamefont {K.}~\bibnamefont {Husain}},
  \bibinfo {author} {\bibfnamefont {J.}~\bibnamefont {Sheng}}, \bibinfo
  {author} {\bibfnamefont {S.}~\bibnamefont {Wang}}, \ and\ \bibinfo {author}
  {\bibfnamefont {A.}~\bibnamefont {Murugan}},\ }\href@noop {} {\bibfield
  {journal} {\bibinfo  {journal} {Proceedings of the National Academy of
  Sciences}\ }\textbf {\bibinfo {volume} {117}},\ \bibinfo {pages} {12693}
  (\bibinfo {year} {2020})}\BibitemShut {NoStop}%
\bibitem [{\citenamefont {Jacot}\ \emph {et~al.}(2018)\citenamefont {Jacot},
  \citenamefont {Gabriel},\ and\ \citenamefont {Hongler}}]{jacot2018neural}%
  \BibitemOpen
  \bibfield  {author} {\bibinfo {author} {\bibfnamefont {A.}~\bibnamefont
  {Jacot}}, \bibinfo {author} {\bibfnamefont {F.}~\bibnamefont {Gabriel}}, \
  and\ \bibinfo {author} {\bibfnamefont {C.}~\bibnamefont {Hongler}},\ }in\
  \href@noop {} {\emph {\bibinfo {booktitle} {Advances in neural information
  processing systems}}}\ (\bibinfo {year} {2018})\ pp.\ \bibinfo {pages}
  {8571--8580}\BibitemShut {NoStop}%
\bibitem [{\citenamefont {Lee}\ \emph {et~al.}(2019)\citenamefont {Lee},
  \citenamefont {Xiao}, \citenamefont {Schoenholz}, \citenamefont {Bahri},
  \citenamefont {Novak}, \citenamefont {Sohl-Dickstein},\ and\ \citenamefont
  {Pennington}}]{lee2019wide}%
  \BibitemOpen
  \bibfield  {author} {\bibinfo {author} {\bibfnamefont {J.}~\bibnamefont
  {Lee}}, \bibinfo {author} {\bibfnamefont {L.}~\bibnamefont {Xiao}}, \bibinfo
  {author} {\bibfnamefont {S.}~\bibnamefont {Schoenholz}}, \bibinfo {author}
  {\bibfnamefont {Y.}~\bibnamefont {Bahri}}, \bibinfo {author} {\bibfnamefont
  {R.}~\bibnamefont {Novak}}, \bibinfo {author} {\bibfnamefont
  {J.}~\bibnamefont {Sohl-Dickstein}}, \ and\ \bibinfo {author} {\bibfnamefont
  {J.}~\bibnamefont {Pennington}},\ }in\ \href@noop {} {\emph {\bibinfo
  {booktitle} {Advances in neural information processing systems}}}\ (\bibinfo
  {year} {2019})\ pp.\ \bibinfo {pages} {8572--8583}\BibitemShut {NoStop}%
\bibitem [{\citenamefont {Zhang}\ \emph {et~al.}(2016)\citenamefont {Zhang},
  \citenamefont {Bengio}, \citenamefont {Hardt}, \citenamefont {Recht},\ and\
  \citenamefont {Vinyals}}]{zhang2016understanding}%
  \BibitemOpen
  \bibfield  {author} {\bibinfo {author} {\bibfnamefont {C.}~\bibnamefont
  {Zhang}}, \bibinfo {author} {\bibfnamefont {S.}~\bibnamefont {Bengio}},
  \bibinfo {author} {\bibfnamefont {M.}~\bibnamefont {Hardt}}, \bibinfo
  {author} {\bibfnamefont {B.}~\bibnamefont {Recht}}, \ and\ \bibinfo {author}
  {\bibfnamefont {O.}~\bibnamefont {Vinyals}},\ }\href@noop {} {\bibfield
  {journal} {\bibinfo  {journal} {arXiv preprint arXiv:1611.03530}\ } (\bibinfo
  {year} {2016})}\BibitemShut {NoStop}%
\bibitem [{\citenamefont {Shankar}\ \emph {et~al.}(2020)\citenamefont
  {Shankar}, \citenamefont {Fang}, \citenamefont {Guo}, \citenamefont
  {Fridovich-Keil}, \citenamefont {Ragan-Kelley}, \citenamefont {Schmidt},\
  and\ \citenamefont {Recht}}]{shankar2020neural}%
  \BibitemOpen
  \bibfield  {author} {\bibinfo {author} {\bibfnamefont {V.}~\bibnamefont
  {Shankar}}, \bibinfo {author} {\bibfnamefont {A.}~\bibnamefont {Fang}},
  \bibinfo {author} {\bibfnamefont {W.}~\bibnamefont {Guo}}, \bibinfo {author}
  {\bibfnamefont {S.}~\bibnamefont {Fridovich-Keil}}, \bibinfo {author}
  {\bibfnamefont {J.}~\bibnamefont {Ragan-Kelley}}, \bibinfo {author}
  {\bibfnamefont {L.}~\bibnamefont {Schmidt}}, \ and\ \bibinfo {author}
  {\bibfnamefont {B.}~\bibnamefont {Recht}},\ }in\ \href@noop {} {\emph
  {\bibinfo {booktitle} {International Conference on Machine Learning}}}\
  (\bibinfo {organization} {PMLR},\ \bibinfo {year} {2020})\ pp.\ \bibinfo
  {pages} {8614--8623}\BibitemShut {NoStop}%
\bibitem [{\citenamefont {Ruiz-Garc{\'\i}a}\ \emph {et~al.}(2019)\citenamefont
  {Ruiz-Garc{\'\i}a}, \citenamefont {Liu},\ and\ \citenamefont
  {Katifori}}]{ruiz2019tuning}%
  \BibitemOpen
  \bibfield  {author} {\bibinfo {author} {\bibfnamefont {M.}~\bibnamefont
  {Ruiz-Garc{\'\i}a}}, \bibinfo {author} {\bibfnamefont {A.~J.}\ \bibnamefont
  {Liu}}, \ and\ \bibinfo {author} {\bibfnamefont {E.}~\bibnamefont
  {Katifori}},\ }\href@noop {} {\bibfield  {journal} {\bibinfo  {journal}
  {Physical Review E}\ }\textbf {\bibinfo {volume} {100}},\ \bibinfo {pages}
  {052608} (\bibinfo {year} {2019})}\BibitemShut {NoStop}%
\bibitem [{\citenamefont {Bradbury}\ \emph {et~al.}(2018)\citenamefont
  {Bradbury}, \citenamefont {Frostig}, \citenamefont {Hawkins}, \citenamefont
  {Johnson}, \citenamefont {Leary}, \citenamefont {Maclaurin}, \citenamefont
  {Necula}, \citenamefont {Paszke}, \citenamefont {Vander{P}las}, \citenamefont
  {Wanderman-{M}ilne},\ and\ \citenamefont {Zhang}}]{jax2018github}%
  \BibitemOpen
  \bibfield  {author} {\bibinfo {author} {\bibfnamefont {J.}~\bibnamefont
  {Bradbury}}, \bibinfo {author} {\bibfnamefont {R.}~\bibnamefont {Frostig}},
  \bibinfo {author} {\bibfnamefont {P.}~\bibnamefont {Hawkins}}, \bibinfo
  {author} {\bibfnamefont {M.~J.}\ \bibnamefont {Johnson}}, \bibinfo {author}
  {\bibfnamefont {C.}~\bibnamefont {Leary}}, \bibinfo {author} {\bibfnamefont
  {D.}~\bibnamefont {Maclaurin}}, \bibinfo {author} {\bibfnamefont
  {G.}~\bibnamefont {Necula}}, \bibinfo {author} {\bibfnamefont
  {A.}~\bibnamefont {Paszke}}, \bibinfo {author} {\bibfnamefont
  {J.}~\bibnamefont {Vander{P}las}}, \bibinfo {author} {\bibfnamefont
  {S.}~\bibnamefont {Wanderman-{M}ilne}}, \ and\ \bibinfo {author}
  {\bibfnamefont {Q.}~\bibnamefont {Zhang}},\ }\href
  {http://github.com/google/jax} {\enquote {\bibinfo {title} {{JAX}: composable
  transformations of {P}ython+{N}um{P}y programs},}\ } (\bibinfo {year}
  {2018})\BibitemShut {NoStop}%
\bibitem [{\citenamefont {Novak}\ \emph {et~al.}(2020)\citenamefont {Novak},
  \citenamefont {Xiao}, \citenamefont {Hron}, \citenamefont {Lee},
  \citenamefont {Alemi}, \citenamefont {Sohl-Dickstein},\ and\ \citenamefont
  {Schoenholz}}]{neuraltangents2020}%
  \BibitemOpen
  \bibfield  {author} {\bibinfo {author} {\bibfnamefont {R.}~\bibnamefont
  {Novak}}, \bibinfo {author} {\bibfnamefont {L.}~\bibnamefont {Xiao}},
  \bibinfo {author} {\bibfnamefont {J.}~\bibnamefont {Hron}}, \bibinfo {author}
  {\bibfnamefont {J.}~\bibnamefont {Lee}}, \bibinfo {author} {\bibfnamefont
  {A.~A.}\ \bibnamefont {Alemi}}, \bibinfo {author} {\bibfnamefont
  {J.}~\bibnamefont {Sohl-Dickstein}}, \ and\ \bibinfo {author} {\bibfnamefont
  {S.~S.}\ \bibnamefont {Schoenholz}},\ }in\ \href@noop {} {\emph {\bibinfo
  {booktitle} {International Conference on Learning Representations}}}\
  (\bibinfo {year} {2020})\BibitemShut {NoStop}%
\bibitem [{\citenamefont {Ghorbani}\ \emph
  {et~al.}(2019{\natexlab{a}})\citenamefont {Ghorbani}, \citenamefont
  {Krishnan},\ and\ \citenamefont {Xiao}}]{ghorbani2019}%
  \BibitemOpen
  \bibfield  {author} {\bibinfo {author} {\bibfnamefont {B.}~\bibnamefont
  {Ghorbani}}, \bibinfo {author} {\bibfnamefont {S.}~\bibnamefont {Krishnan}},
  \ and\ \bibinfo {author} {\bibfnamefont {Y.}~\bibnamefont {Xiao}},\
  }\href@noop {} {\bibfield  {journal} {\bibinfo  {journal} {CoRR}\ }\textbf
  {\bibinfo {volume} {abs/1901.10159}} (\bibinfo {year}
  {2019}{\natexlab{a}})},\ \Eprint {http://arxiv.org/abs/1901.10159}
  {1901.10159} \BibitemShut {NoStop}%
\bibitem [{\citenamefont {Zagoruyko}\ and\ \citenamefont
  {Komodakis}(2016)}]{wrn}%
  \BibitemOpen
  \bibfield  {author} {\bibinfo {author} {\bibfnamefont {S.}~\bibnamefont
  {Zagoruyko}}\ and\ \bibinfo {author} {\bibfnamefont {N.}~\bibnamefont
  {Komodakis}},\ }\href {http://arxiv.org/abs/1605.07146} {\bibfield  {journal}
  {\bibinfo  {journal} {CoRR}\ }\textbf {\bibinfo {volume} {abs/1605.07146}}
  (\bibinfo {year} {2016})},\ \Eprint {http://arxiv.org/abs/1605.07146}
  {arXiv:1605.07146} \BibitemShut {NoStop}%
\bibitem [{\citenamefont {Karpathy}\ \emph {et~al.}(2020)\citenamefont
  {Karpathy} \emph {et~al.}}]{karpathy2020convolutional}%
  \BibitemOpen
  \bibfield  {author} {\bibinfo {author} {\bibfnamefont {A.}~\bibnamefont
  {Karpathy}} \emph {et~al.},\ }\href@noop {} {\bibfield  {journal} {\bibinfo
  {journal} {Course notes hosted on GitHub. Retrieved from: http://cs231n.
  github. io}\ } (\bibinfo {year} {2020})}\BibitemShut {NoStop}%
\bibitem [{\citenamefont {Ghorbani}\ \emph
  {et~al.}(2019{\natexlab{b}})\citenamefont {Ghorbani}, \citenamefont
  {Krishnan},\ and\ \citenamefont {Xiao}}]{ghorbani2019investigation}%
  \BibitemOpen
  \bibfield  {author} {\bibinfo {author} {\bibfnamefont {B.}~\bibnamefont
  {Ghorbani}}, \bibinfo {author} {\bibfnamefont {S.}~\bibnamefont {Krishnan}},
  \ and\ \bibinfo {author} {\bibfnamefont {Y.}~\bibnamefont {Xiao}},\
  }\href@noop {} {\bibfield  {journal} {\bibinfo  {journal} {arXiv preprint
  arXiv:1901.10159}\ } (\bibinfo {year} {2019}{\natexlab{b}})}\BibitemShut
  {NoStop}%
\bibitem [{\citenamefont {Gilmer}(2020)}]{gilmer2020}%
  \BibitemOpen
  \bibfield  {author} {\bibinfo {author} {\bibfnamefont {J.}~\bibnamefont
  {Gilmer}},\ }\href@noop {} {\enquote {\bibinfo {title} {Large scale spectral
  density estimation for deep neural networks},}\ }\bibinfo {howpublished}
  {\url{https://github.com/google/spectral-density}} (\bibinfo {year}
  {2020})\BibitemShut {NoStop}%
\bibitem [{\citenamefont {Sagun}\ \emph {et~al.}(2017)\citenamefont {Sagun},
  \citenamefont {Evci}, \citenamefont {Guney}, \citenamefont {Dauphin},\ and\
  \citenamefont {Bottou}}]{sagun2017empirical}%
  \BibitemOpen
  \bibfield  {author} {\bibinfo {author} {\bibfnamefont {L.}~\bibnamefont
  {Sagun}}, \bibinfo {author} {\bibfnamefont {U.}~\bibnamefont {Evci}},
  \bibinfo {author} {\bibfnamefont {V.~U.}\ \bibnamefont {Guney}}, \bibinfo
  {author} {\bibfnamefont {Y.}~\bibnamefont {Dauphin}}, \ and\ \bibinfo
  {author} {\bibfnamefont {L.}~\bibnamefont {Bottou}},\ }\href@noop {}
  {\bibfield  {journal} {\bibinfo  {journal} {arXiv preprint arXiv:1706.04454}\
  } (\bibinfo {year} {2017})}\BibitemShut {NoStop}%
\bibitem [{\citenamefont {Sagun}\ \emph {et~al.}(2016)\citenamefont {Sagun},
  \citenamefont {Bottou},\ and\ \citenamefont {LeCun}}]{sagun2016eigenvalues}%
  \BibitemOpen
  \bibfield  {author} {\bibinfo {author} {\bibfnamefont {L.}~\bibnamefont
  {Sagun}}, \bibinfo {author} {\bibfnamefont {L.}~\bibnamefont {Bottou}}, \
  and\ \bibinfo {author} {\bibfnamefont {Y.}~\bibnamefont {LeCun}},\
  }\href@noop {} {\bibfield  {journal} {\bibinfo  {journal} {arXiv preprint
  arXiv:1611.07476}\ } (\bibinfo {year} {2016})}\BibitemShut {NoStop}%
\bibitem [{\citenamefont {Gur-Ari}\ \emph {et~al.}(2018)\citenamefont
  {Gur-Ari}, \citenamefont {Roberts},\ and\ \citenamefont
  {Dyer}}]{gur2018gradient}%
  \BibitemOpen
  \bibfield  {author} {\bibinfo {author} {\bibfnamefont {G.}~\bibnamefont
  {Gur-Ari}}, \bibinfo {author} {\bibfnamefont {D.~A.}\ \bibnamefont
  {Roberts}}, \ and\ \bibinfo {author} {\bibfnamefont {E.}~\bibnamefont
  {Dyer}},\ }\href@noop {} {\bibfield  {journal} {\bibinfo  {journal} {arXiv
  preprint arXiv:1812.04754}\ } (\bibinfo {year} {2018})}\BibitemShut {NoStop}%
\bibitem [{\citenamefont {Lewkowycz}\ \emph {et~al.}(2020)\citenamefont
  {Lewkowycz}, \citenamefont {Bahri}, \citenamefont {Dyer}, \citenamefont
  {Sohl-Dickstein},\ and\ \citenamefont {Gur-Ari}}]{lewkowycz2020large}%
  \BibitemOpen
  \bibfield  {author} {\bibinfo {author} {\bibfnamefont {A.}~\bibnamefont
  {Lewkowycz}}, \bibinfo {author} {\bibfnamefont {Y.}~\bibnamefont {Bahri}},
  \bibinfo {author} {\bibfnamefont {E.}~\bibnamefont {Dyer}}, \bibinfo {author}
  {\bibfnamefont {J.}~\bibnamefont {Sohl-Dickstein}}, \ and\ \bibinfo {author}
  {\bibfnamefont {G.}~\bibnamefont {Gur-Ari}},\ }\href@noop {} {\bibfield
  {journal} {\bibinfo  {journal} {arXiv preprint arXiv:2003.02218}\ } (\bibinfo
  {year} {2020})}\BibitemShut {NoStop}%
\bibitem [{\citenamefont {Le~Cun}\ \emph {et~al.}(1991)\citenamefont {Le~Cun},
  \citenamefont {Kanter},\ and\ \citenamefont {Solla}}]{le1991eigenvalues}%
  \BibitemOpen
  \bibfield  {author} {\bibinfo {author} {\bibfnamefont {Y.}~\bibnamefont
  {Le~Cun}}, \bibinfo {author} {\bibfnamefont {I.}~\bibnamefont {Kanter}}, \
  and\ \bibinfo {author} {\bibfnamefont {S.~A.}\ \bibnamefont {Solla}},\
  }\href@noop {} {\bibfield  {journal} {\bibinfo  {journal} {Physical Review
  Letters}\ }\textbf {\bibinfo {volume} {66}},\ \bibinfo {pages} {2396}
  (\bibinfo {year} {1991})}\BibitemShut {NoStop}%
\bibitem [{\citenamefont {Yang}\ and\ \citenamefont
  {Littwin}(2021)}]{yang2021tensor}%
  \BibitemOpen
  \bibfield  {author} {\bibinfo {author} {\bibfnamefont {G.}~\bibnamefont
  {Yang}}\ and\ \bibinfo {author} {\bibfnamefont {E.}~\bibnamefont {Littwin}},\
  }\href@noop {} {\enquote {\bibinfo {title} {Tensor programs iib:
  Architectural universality of neural tangent kernel training dynamics},}\ }
  (\bibinfo {year} {2021}),\ \Eprint {http://arxiv.org/abs/2105.03703}
  {arXiv:2105.03703 [cs.LG]} \BibitemShut {NoStop}%
\bibitem [{\citenamefont {Xiao}\ \emph {et~al.}(2020)\citenamefont {Xiao},
  \citenamefont {Pennington},\ and\ \citenamefont {Schoenholz}}]{xiao2020}%
  \BibitemOpen
  \bibfield  {author} {\bibinfo {author} {\bibfnamefont {L.}~\bibnamefont
  {Xiao}}, \bibinfo {author} {\bibfnamefont {J.}~\bibnamefont {Pennington}}, \
  and\ \bibinfo {author} {\bibfnamefont {S.}~\bibnamefont {Schoenholz}},\ }in\
  \href@noop {} {\emph {\bibinfo {booktitle} {Proceedings of the 37th
  International Conference on Machine Learning}}},\ \bibinfo {series}
  {Proceedings of Machine Learning Research}, Vol.\ \bibinfo {volume} {119},\
  \bibinfo {editor} {edited by\ \bibinfo {editor} {\bibfnamefont {H.~D.}\
  \bibnamefont {III}}\ and\ \bibinfo {editor} {\bibfnamefont {A.}~\bibnamefont
  {Singh}}}\ (\bibinfo  {publisher} {PMLR},\ \bibinfo {year} {2020})\ pp.\
  \bibinfo {pages} {10462--10472}\BibitemShut {NoStop}%
\bibitem [{\citenamefont {Dauphin}\ and\ \citenamefont
  {Schoenholz}(2019)}]{dauphin2019}%
  \BibitemOpen
  \bibfield  {author} {\bibinfo {author} {\bibfnamefont {Y.~N.}\ \bibnamefont
  {Dauphin}}\ and\ \bibinfo {author} {\bibfnamefont {S.}~\bibnamefont
  {Schoenholz}},\ }in\ \href@noop {} {\emph {\bibinfo {booktitle} {Advances in
  Neural Information Processing Systems}}},\ Vol.~\bibinfo {volume} {32},\
  \bibinfo {editor} {edited by\ \bibinfo {editor} {\bibfnamefont
  {H.}~\bibnamefont {Wallach}}, \bibinfo {editor} {\bibfnamefont
  {H.}~\bibnamefont {Larochelle}}, \bibinfo {editor} {\bibfnamefont
  {A.}~\bibnamefont {Beygelzimer}}, \bibinfo {editor} {\bibfnamefont
  {F.}~\bibnamefont {d'~Alch\'{e}-Buc}}, \bibinfo {editor} {\bibfnamefont
  {E.}~\bibnamefont {Fox}}, \ and\ \bibinfo {editor} {\bibfnamefont
  {R.}~\bibnamefont {Garnett}}}\ (\bibinfo  {publisher} {Curran Associates,
  Inc.},\ \bibinfo {year} {2019})\BibitemShut {NoStop}%
\bibitem [{\citenamefont {Jacot}\ \emph {et~al.}(2019)\citenamefont {Jacot},
  \citenamefont {Gabriel},\ and\ \citenamefont {Hongler}}]{jacot2019}%
  \BibitemOpen
  \bibfield  {author} {\bibinfo {author} {\bibfnamefont {A.}~\bibnamefont
  {Jacot}}, \bibinfo {author} {\bibfnamefont {F.}~\bibnamefont {Gabriel}}, \
  and\ \bibinfo {author} {\bibfnamefont {C.}~\bibnamefont {Hongler}},\ }\href
  {http://arxiv.org/abs/1907.05715} {\bibfield  {journal} {\bibinfo  {journal}
  {CoRR}\ }\textbf {\bibinfo {volume} {abs/1907.05715}} (\bibinfo {year}
  {2019})},\ \Eprint {http://arxiv.org/abs/1907.05715} {arXiv:1907.05715}
  \BibitemShut {NoStop}%
\bibitem [{\citenamefont {Agarwala}\ \emph {et~al.}(2020)\citenamefont
  {Agarwala}, \citenamefont {Pennington}, \citenamefont {Dauphin},\ and\
  \citenamefont {Schoenholz}}]{agarwala2020}%
  \BibitemOpen
  \bibfield  {author} {\bibinfo {author} {\bibfnamefont {A.}~\bibnamefont
  {Agarwala}}, \bibinfo {author} {\bibfnamefont {J.}~\bibnamefont
  {Pennington}}, \bibinfo {author} {\bibfnamefont {Y.~N.}\ \bibnamefont
  {Dauphin}}, \ and\ \bibinfo {author} {\bibfnamefont {S.~S.}\ \bibnamefont
  {Schoenholz}},\ }\href@noop {} {\bibfield  {journal} {\bibinfo  {journal}
  {CoRR}\ }\textbf {\bibinfo {volume} {abs/2010.07344}} (\bibinfo {year}
  {2020})},\ \Eprint {http://arxiv.org/abs/2010.07344} {arXiv:2010.07344}
  \BibitemShut {NoStop}%
\bibitem [{\citenamefont {Goodfellow}\ \emph {et~al.}(2013)\citenamefont
  {Goodfellow}, \citenamefont {Mirza}, \citenamefont {Xiao}, \citenamefont
  {Courville},\ and\ \citenamefont {Bengio}}]{goodfellow2013empirical}%
  \BibitemOpen
  \bibfield  {author} {\bibinfo {author} {\bibfnamefont {I.~J.}\ \bibnamefont
  {Goodfellow}}, \bibinfo {author} {\bibfnamefont {M.}~\bibnamefont {Mirza}},
  \bibinfo {author} {\bibfnamefont {D.}~\bibnamefont {Xiao}}, \bibinfo {author}
  {\bibfnamefont {A.}~\bibnamefont {Courville}}, \ and\ \bibinfo {author}
  {\bibfnamefont {Y.}~\bibnamefont {Bengio}},\ }\href@noop {} {\bibfield
  {journal} {\bibinfo  {journal} {arXiv preprint arXiv:1312.6211}\ } (\bibinfo
  {year} {2013})}\BibitemShut {NoStop}%
\bibitem [{\citenamefont {Ratcliff}(1990)}]{ratcliff1990connectionist}%
  \BibitemOpen
  \bibfield  {author} {\bibinfo {author} {\bibfnamefont {R.}~\bibnamefont
  {Ratcliff}},\ }\href@noop {} {\bibfield  {journal} {\bibinfo  {journal}
  {Psychological review}\ }\textbf {\bibinfo {volume} {97}},\ \bibinfo {pages}
  {285} (\bibinfo {year} {1990})}\BibitemShut {NoStop}%
\bibitem [{\citenamefont {McCloskey}\ and\ \citenamefont
  {Cohen}(1989)}]{mccloskey1989catastrophic}%
  \BibitemOpen
  \bibfield  {author} {\bibinfo {author} {\bibfnamefont {M.}~\bibnamefont
  {McCloskey}}\ and\ \bibinfo {author} {\bibfnamefont {N.~J.}\ \bibnamefont
  {Cohen}},\ }in\ \href@noop {} {\emph {\bibinfo {booktitle} {Psychology of
  learning and motivation}}},\ Vol.~\bibinfo {volume} {24}\ (\bibinfo
  {publisher} {Elsevier},\ \bibinfo {year} {1989})\ pp.\ \bibinfo {pages}
  {109--165}\BibitemShut {NoStop}%
\bibitem [{\citenamefont {Towns}\ \emph {et~al.}(2014)\citenamefont {Towns},
  \citenamefont {Cockerill}, \citenamefont {Dahan}, \citenamefont {Foster},
  \citenamefont {Gaither}, \citenamefont {Grimshaw}, \citenamefont {Hazlewood},
  \citenamefont {Lathrop}, \citenamefont {Lifka}, \citenamefont {Peterson}
  \emph {et~al.}}]{towns2014xsede}%
  \BibitemOpen
  \bibfield  {author} {\bibinfo {author} {\bibfnamefont {J.}~\bibnamefont
  {Towns}}, \bibinfo {author} {\bibfnamefont {T.}~\bibnamefont {Cockerill}},
  \bibinfo {author} {\bibfnamefont {M.}~\bibnamefont {Dahan}}, \bibinfo
  {author} {\bibfnamefont {I.}~\bibnamefont {Foster}}, \bibinfo {author}
  {\bibfnamefont {K.}~\bibnamefont {Gaither}}, \bibinfo {author} {\bibfnamefont
  {A.}~\bibnamefont {Grimshaw}}, \bibinfo {author} {\bibfnamefont
  {V.}~\bibnamefont {Hazlewood}}, \bibinfo {author} {\bibfnamefont
  {S.}~\bibnamefont {Lathrop}}, \bibinfo {author} {\bibfnamefont
  {D.}~\bibnamefont {Lifka}}, \bibinfo {author} {\bibfnamefont {G.~D.}\
  \bibnamefont {Peterson}},  \emph {et~al.},\ }\href@noop {} {\bibfield
  {journal} {\bibinfo  {journal} {Computing in science \& engineering}\
  }\textbf {\bibinfo {volume} {16}},\ \bibinfo {pages} {62} (\bibinfo {year}
  {2014})}\BibitemShut {NoStop}%
\end{thebibliography}%


\begin{thebibliography}{4}%
\makeatletter
\providecommand \@ifxundefined [1]{%
 \@ifx{#1\undefined}
}%
\providecommand \@ifnum [1]{%
 \ifnum #1\expandafter \@firstoftwo
 \else \expandafter \@secondoftwo
 \fi
}%
\providecommand \@ifx [1]{%
 \ifx #1\expandafter \@firstoftwo
 \else \expandafter \@secondoftwo
 \fi
}%
\providecommand \natexlab [1]{#1}%
\providecommand \enquote  [1]{``#1''}%
\providecommand \bibnamefont  [1]{#1}%
\providecommand \bibfnamefont [1]{#1}%
\providecommand \citenamefont [1]{#1}%
\providecommand \href@noop [0]{\@secondoftwo}%
\providecommand \href [0]{\begingroup \@sanitize@url \@href}%
\providecommand \@href[1]{\@@startlink{#1}\@@href}%
\providecommand \@@href[1]{\endgroup#1\@@endlink}%
\providecommand \@sanitize@url [0]{\catcode `\\12\catcode `\$12\catcode
  `\&12\catcode `\#12\catcode `\^12\catcode `\_12\catcode `\%12\relax}%
\providecommand \@@startlink[1]{}%
\providecommand \@@endlink[0]{}%
\providecommand \url  [0]{\begingroup\@sanitize@url \@url }%
\providecommand \@url [1]{\endgroup\@href {#1}{\urlprefix }}%
\providecommand \urlprefix  [0]{URL }%
\providecommand \Eprint [0]{\href }%
\providecommand \doibase [0]{http://dx.doi.org/}%
\providecommand \selectlanguage [0]{\@gobble}%
\providecommand \bibinfo  [0]{\@secondoftwo}%
\providecommand \bibfield  [0]{\@secondoftwo}%
\providecommand \translation [1]{[#1]}%
\providecommand \BibitemOpen [0]{}%
\providecommand \bibitemStop [0]{}%
\providecommand \bibitemNoStop [0]{.\EOS\space}%
\providecommand \EOS [0]{\spacefactor3000\relax}%
\providecommand \BibitemShut  [1]{\csname bibitem#1\endcsname}%
\let\auto@bib@innerbib\@empty
\bibitem [{\citenamefont {Ghorbani}\ \emph {et~al.}(2019)\citenamefont
  {Ghorbani}, \citenamefont {Krishnan},\ and\ \citenamefont
  {Xiao}}]{ghorbani2019investigation}%
  \BibitemOpen
  \bibfield  {author} {\bibinfo {author} {\bibfnamefont {B.}~\bibnamefont
  {Ghorbani}}, \bibinfo {author} {\bibfnamefont {S.}~\bibnamefont {Krishnan}},
  \ and\ \bibinfo {author} {\bibfnamefont {Y.}~\bibnamefont {Xiao}},\
  }\href@noop {} {\bibfield  {journal} {\bibinfo  {journal} {arXiv preprint
  arXiv:1901.10159}\ } (\bibinfo {year} {2019})}\BibitemShut {NoStop}%
\bibitem [{\citenamefont {Gilmer}(2020)}]{gilmer2020}%
  \BibitemOpen
  \bibfield  {author} {\bibinfo {author} {\bibfnamefont {J.}~\bibnamefont
  {Gilmer}},\ }\href@noop {} {\enquote {\bibinfo {title} {Large scale spectral
  density estimation for deep neural networks},}\ }\bibinfo {howpublished}
  {\url{https://github.com/google/spectral-density}} (\bibinfo {year}
  {2020})\BibitemShut {NoStop}%
\bibitem [{\citenamefont {Shankar}\ \emph {et~al.}(2020)\citenamefont
  {Shankar}, \citenamefont {Fang}, \citenamefont {Guo}, \citenamefont
  {Fridovich-Keil}, \citenamefont {Ragan-Kelley}, \citenamefont {Schmidt},\
  and\ \citenamefont {Recht}}]{shankar2020neural}%
  \BibitemOpen
  \bibfield  {author} {\bibinfo {author} {\bibfnamefont {V.}~\bibnamefont
  {Shankar}}, \bibinfo {author} {\bibfnamefont {A.}~\bibnamefont {Fang}},
  \bibinfo {author} {\bibfnamefont {W.}~\bibnamefont {Guo}}, \bibinfo {author}
  {\bibfnamefont {S.}~\bibnamefont {Fridovich-Keil}}, \bibinfo {author}
  {\bibfnamefont {J.}~\bibnamefont {Ragan-Kelley}}, \bibinfo {author}
  {\bibfnamefont {L.}~\bibnamefont {Schmidt}}, \ and\ \bibinfo {author}
  {\bibfnamefont {B.}~\bibnamefont {Recht}},\ }in\ \href@noop {} {\emph
  {\bibinfo {booktitle} {International Conference on Machine Learning}}}\
  (\bibinfo {organization} {PMLR},\ \bibinfo {year} {2020})\ pp.\ \bibinfo
  {pages} {8614--8623}\BibitemShut {NoStop}%
\bibitem [{\citenamefont {Zagoruyko}\ and\ \citenamefont
  {Komodakis}(2016)}]{wrn}%
  \BibitemOpen
  \bibfield  {author} {\bibinfo {author} {\bibfnamefont {S.}~\bibnamefont
  {Zagoruyko}}\ and\ \bibinfo {author} {\bibfnamefont {N.}~\bibnamefont
  {Komodakis}},\ }\href {http://arxiv.org/abs/1605.07146} {\bibfield  {journal}
  {\bibinfo  {journal} {CoRR}\ }\textbf {\bibinfo {volume} {abs/1605.07146}}
  (\bibinfo {year} {2016})},\ \Eprint {http://arxiv.org/abs/1605.07146}
  {arXiv:1605.07146} \BibitemShut {NoStop}%
\end{thebibliography}%

\end{document}


\title{Supplementary Materials for ``Tilting the playing field: Dynamical loss functions for machine learning''}

\author{Miguel Ruiz-Garc\'ia}

\affiliation{Department of Physics and Astronomy, University of Pennsylvania, Philadelphia, PA 19104, USA}

\affiliation{Department of Applied Mathematics, ETSII, Universidad Polit\'ecnica de Madrid, Madrid, Spain}

\author{Ge Zhang}

\affiliation{Department of Physics and Astronomy, University of Pennsylvania, Philadelphia, PA 19104, USA}

\author{Samuel S. Schoenholz}

\affiliation{Google Research: Brain Team}

\author{Andrea J. Liu}

\affiliation{Department of Physics and Astronomy, University of Pennsylvania, Philadelphia, PA 19104, USA}

\date{\today}

\maketitle

\section{Second and third largest eigenvalues of the Hessian trigger subsequent bifurcations.}

\begin{figure}
    \centering
    \includegraphics[width=\linewidth]{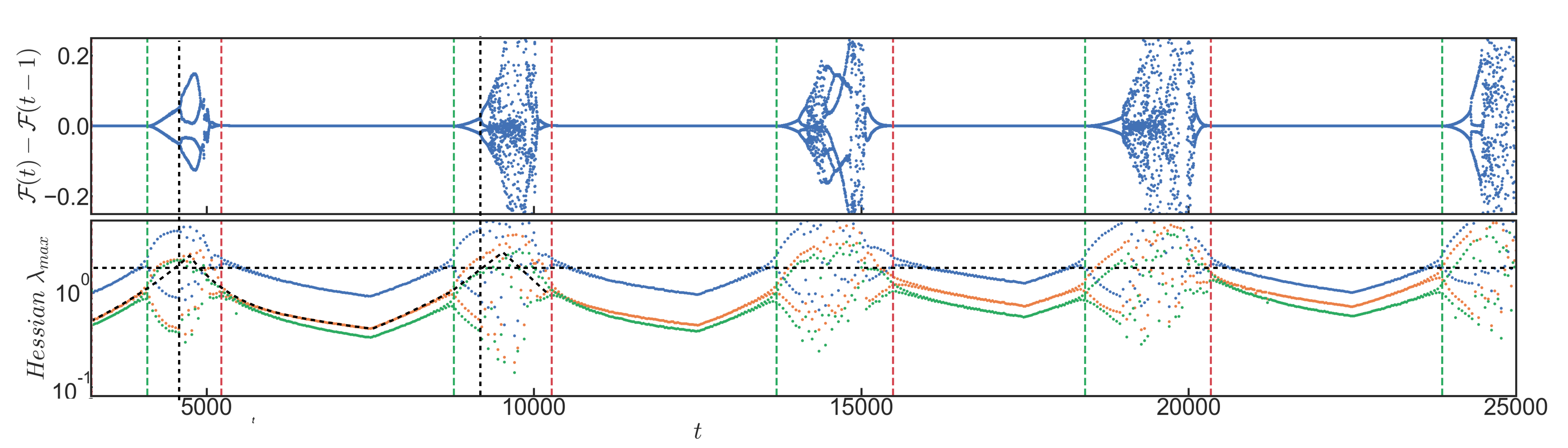}
    \caption{Tracking the three largest eigenvalues of the Hessian. As in figure 4 of the main text, we use a neural network whose  hidden layer has $100$ units, $T=5000$ and $A=70$. Upper panel shows $\mathcal{F}(t)-\mathcal{F}(t-1)$ to display the instabilities more clearly. Bottom panel shows the three largest eigenvalues of the Hessian of the loss function computed using the Lanczos algorithm as described in \cite{ghorbani2019investigation} (we have used an implementation in Google-JAX \cite{gilmer2020}).  Vertical green and red dashed lines mark the times at which Hessian $\lambda_{max}(t) - \lambda_{max}(t-1) \sim 0.1$ corresponding to the start and finish of the instabilities. Averaging Hessian $\lambda_{max}$ at these times we get the horizontal dashed line in the bottom panel, the threshold above which instabilities occur. We have included a new dashed line approximately following the second largest eigenvalue (orange line). Subsequent bifurcations seem to occur when smaller eigenvalues cross the same threshold.
    }
    \label{fig_1}
\end{figure}

We replot here two panels of Fig. 4 of the main text, see Fig. \ref{fig_1}. It shows the behavior of the system as it descends in the dynamical loss function landscape. We use the  spiral  dataset  for  a  case  with  a  rather  high  period  of $T= 5000$ minimization steps and amplitude of $A= 70$, chosen for ease of visualization. In Fig. \ref{fig_1} we include the three largest eigenvalues of the Hessian instead of only one. Indeed, second and third bifurcations seem to correspond to the second and third largest eigenvalues crossing \underline{the same} threshold. Since all eigenvalues are also affected by the bifurcations, we have included new dashed lines to guide the eye.

\section{Hessian eigenvalues also trigger bifurcations using Myrtle5 and CIFAR10}

\begin{figure}
    \centering
    \includegraphics[width=\linewidth]{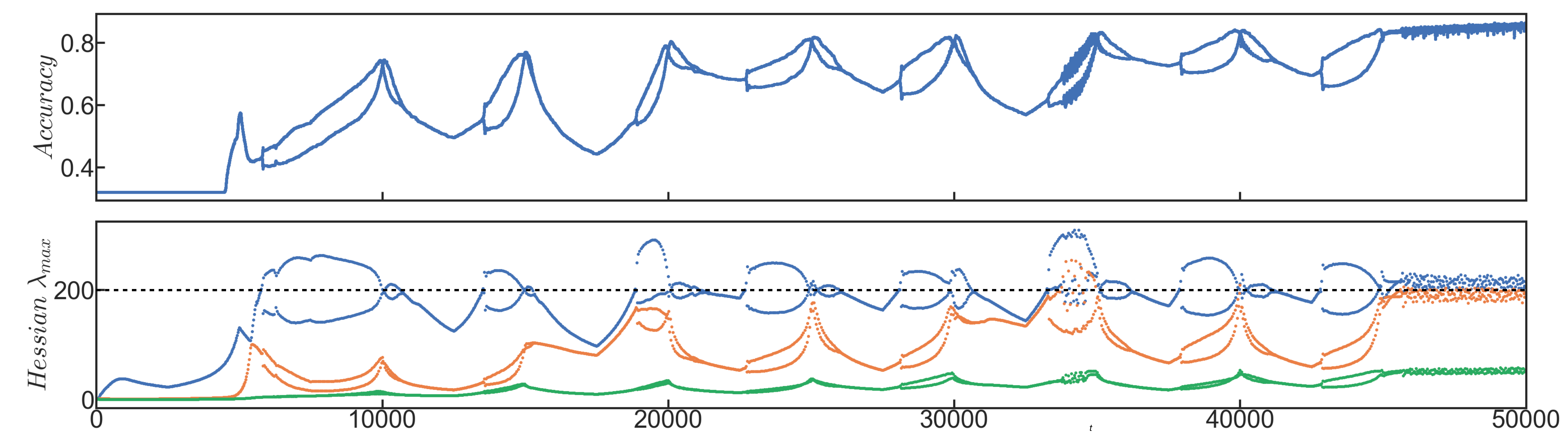}
    \caption{Bifurcations correspond to eigenvalues of the Hessian crossing a threshold when using Myrtle5 to classify a subset of CIFAR10. For simplicity we use a subset of CIFAR10 with $3000$ samples of the first three classes. Bifurcations are clearly present in the Accuracy and they occur when the largest eigenvalue of the Hessian crosses a threshold marked with a dashed line.}
    \label{fig_2}
\end{figure}

In the main text we have studied in detail the dynamics of learning with dynamical loss functions using the spiral dataset because the size of the model and the dataset made it computationally much cheaper, however, bifurcations are a general effect that is also present when using other models and datasets. To show this, we include here an example using Myrtle5 to classify CIFAR10 with a dynamical loss function as the one described in the main text. To make the training dataset more tractable we use a subset of $3000$ samples of the first 3 classes of CIFAR10. Fig. \ref{fig_2} shows that the dynamics follow the same phenomenology described in detail with the spiral dataset in the main text. The accuracy presents bifurcations analogous to Fig. 4 in the main text, and they occur when the largest eigenvalues of the Hessian cross a threshold.

\section{Stopping the oscillations at the end of training}

\begin{figure}
    \centering
    \includegraphics[width=0.65\linewidth]{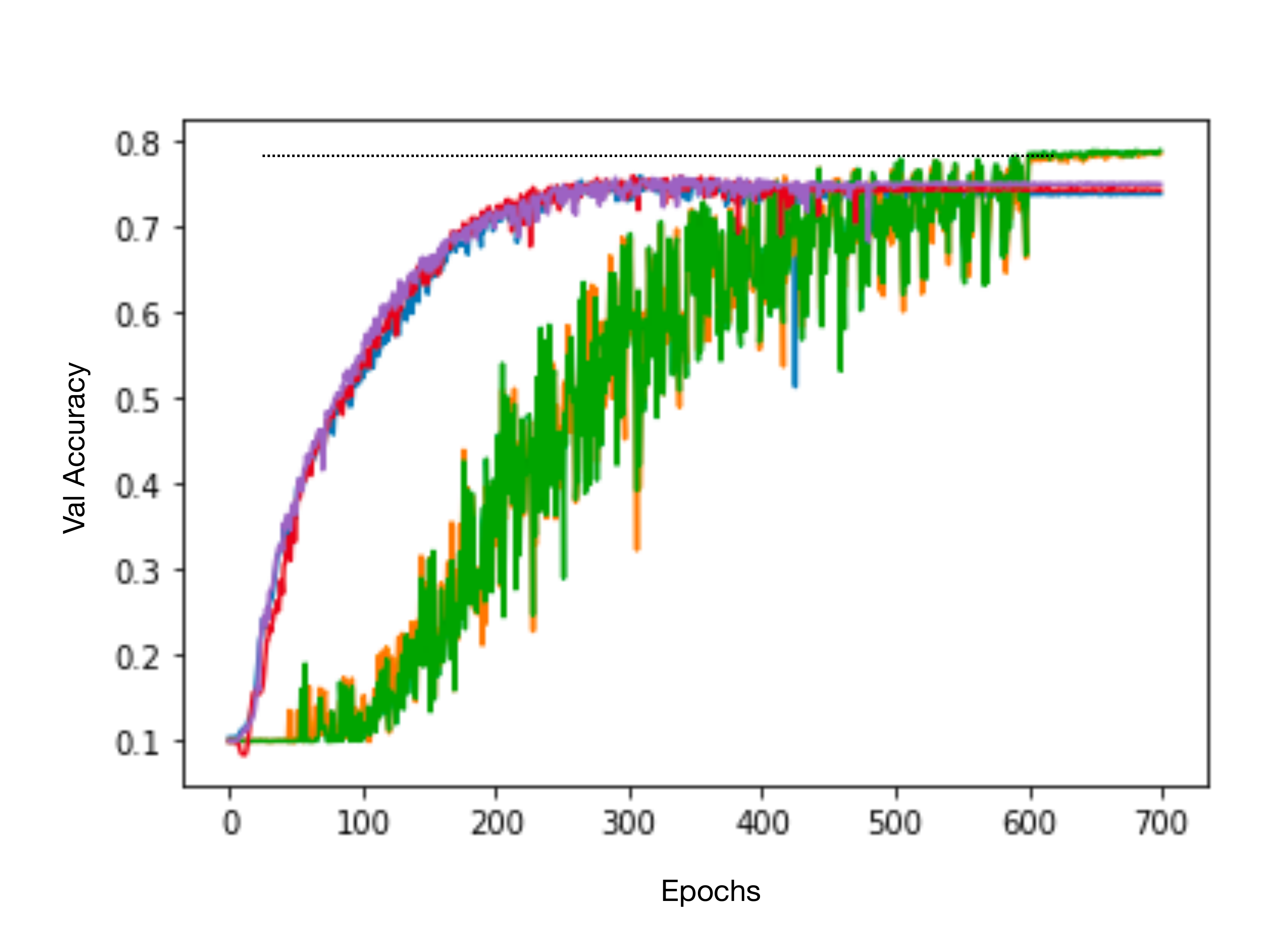}
    \caption{Validation accuracy during the training of Myrtle5 on CIFAR10, as shown in the phase diagrams of Fig. $1$ of the main text. The lines that rise very fast followed by a moderate over-fitting correspond to learning without oscillations. The other group of lines correspond to learning in the region $A\sim 50$ and $T\sim 100$. We stopped the oscillations after epoch $600$. We have included a black dotted line to guide the eye, the final accuracy of the simulations with the oscillations is higher than any point belonging to the simulations without oscillations.  }
    \label{fig_3}
\end{figure}

Fig. 1 of the main text shows two phase diagrams for the dynamical loss function applied to Myrtle5 and CIFAR10. This neural network was adapted from \cite{shankar2020neural}. We used $64$ channels, Nesterov optimizer with momentum $=0.9$, minibatch size $512$, a linear learning rate schedule starting at $0$, reaching $0.02$ in the epoch $300$ and decreasing to $0.002$ in the final epoch ($700$).
Fig. 1 of the main text showed that using the standard cross entropy loss function without the oscillations ($\Gamma_i=1$, $A=1$ line in both panels) the system was able to fit all the training data (training accuracy $\sim 1$) and achieved a $\sim 0.73$ validation accuracy. However, the validation accuracy improved up to $6\%$ thanks to the oscillations for $A\sim 50$ and $T\sim 100$. 
For all $A$ and $T$ the oscillations stopped at epoch $600$ (we changed $A=1$ at that point for all the simulations). Even when the learning rate is decreasing in the second part of each simulation, the oscillations did not reduce appreciably in size at the end of training. In the last $100$ epochs of training, the learning rate is already close to $0.002$, but making $A=1$ (removing the oscillations) helped the system to stabilize and increase the validation accuracy.
We have included here Fig. \ref{fig_3} where we plot two groups of simulations. One group corresponds to learning without oscillations and the other one corresponds to the point in the $(T,A)$ region where validation accuracy improved the most. Learning with oscillations is slower in this case but it reaches a higher validation accuracy. We have included a black dotted line to guide the eye. We leave a systematic study for future work, but at least in this case, stopping the oscillations in the last part of learning had a positive effect, helping the model to achieve a higher validation accuracy.

\section{Wide Residual Networks}

In addition to the Myrtle5 architecture, we also ran a number of experiments on Wide Residual Networks~\cite{wrn}. We used a standard 28-10 residual network with batch normalization and ReLU activation functions. We trained for 200 epochs using a batch size of 1024 running on a 2x2 TPUv2 with a batch size of 128 per chip. We used cosine learning rate decay with an initial learning rate of 0.1 along with the momentum optimizer. Finally we used augmented the data using random flips and crops. This model gets slightly lower accuracy than the version in the literature (95.5$\%$ vs 95.8$\%$) due to the larger batch size employed here. For this architecture we do not observe a statistically significant improvement to the test performance by using an oscillatory loss. 

\begin{figure}
    \centering
    \includegraphics[width=0.45\linewidth]{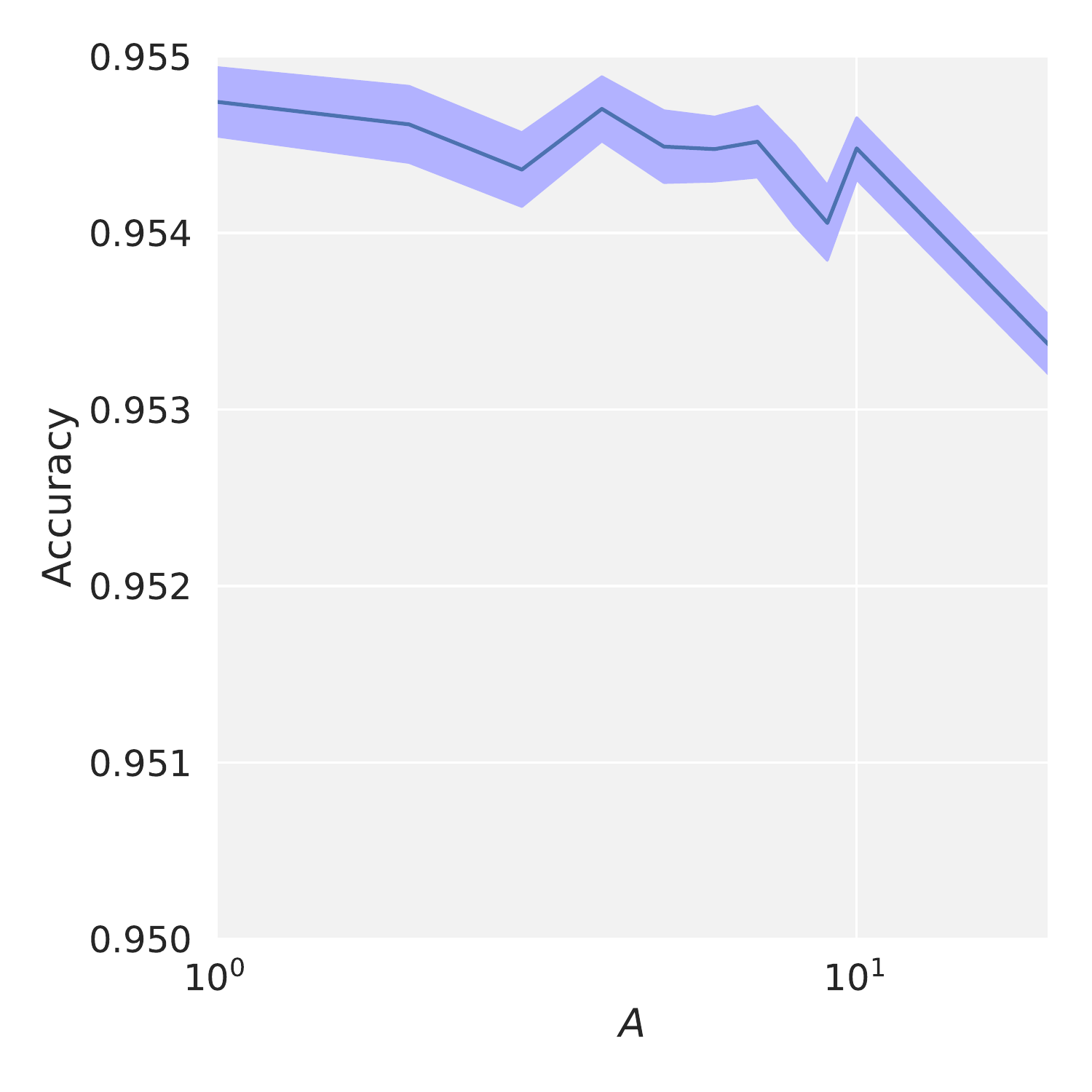}
    \includegraphics[width=0.45\linewidth]{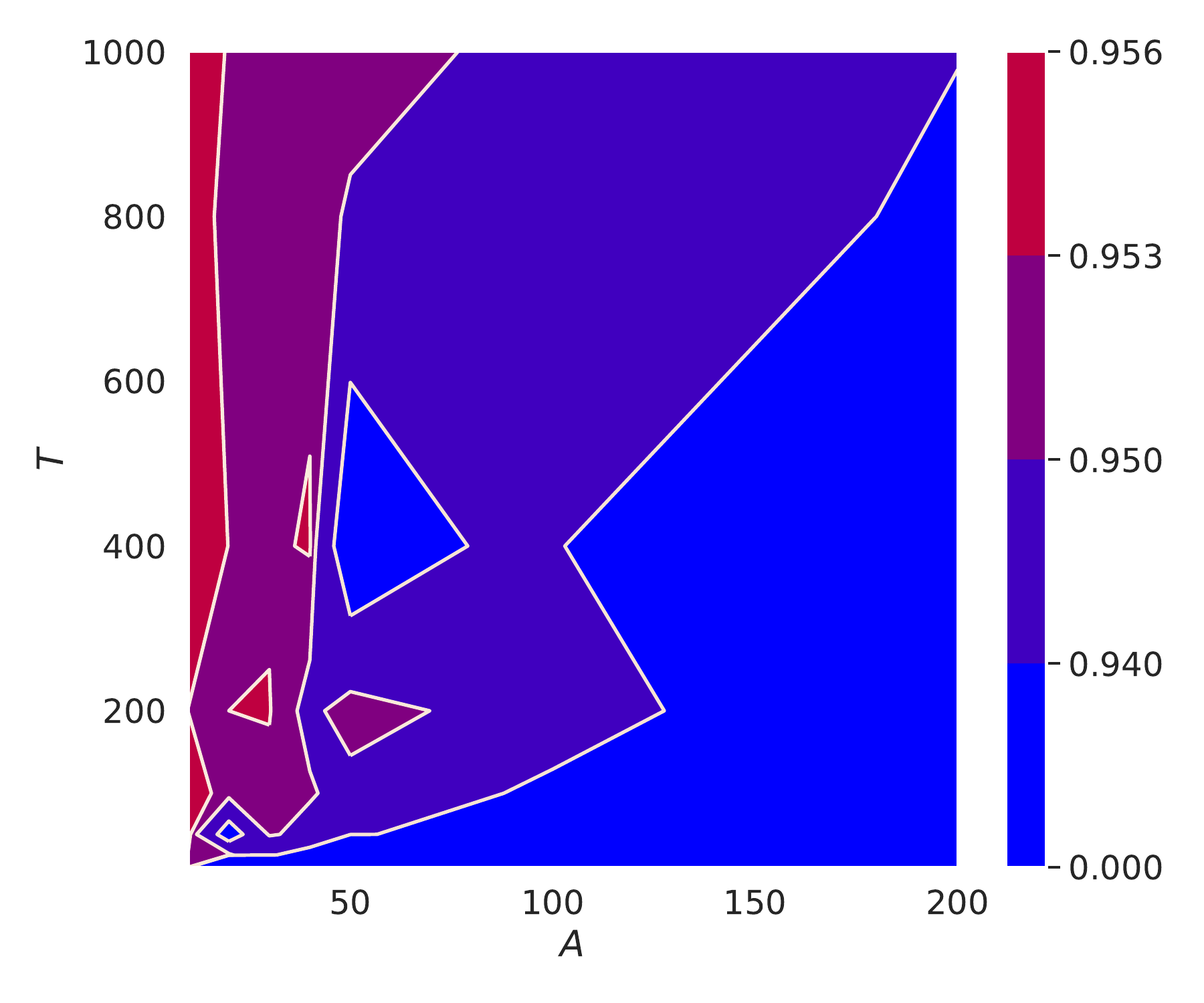}
    \caption{Test accuracy of a Wide Residual Network. Left: The accuracy as a function of oscillation amplitude over a one-dimensional slice with $T = 200$ averaged over 50 seeds. The shaded region shows one standard deviation of the average test accuracy after training. Right: A two dimensional phase plot showing test accuracy as a function of both period and amplitude. }
    \label{fig_3}
\end{figure}


\bibliography{dyn_loss_bib.bib}